\documentclass[journal]{IEEEtran}
\usepackage{cite}

\ifCLASSINFOpdf
  \usepackage[pdftex]{graphicx}
  \graphicspath{{figures/},{biography/}}
\else
  \usepackage[dvips]{graphicx}
  \graphicspath{{figures/},{biography/}}
\fi
\usepackage{amsmath}
\ifCLASSOPTIONcompsoc
 \usepackage[caption=false,font=normalsize,labelfont=sf,textfont=sf]{subfig}
\else
 \usepackage[caption=false,font=footnotesize]{subfig}
\fi
\usepackage{url}

\usepackage{color,colortbl}

\begin{document}
%
\title{Bi-stream Pose Guided Region Ensemble Network for Fingertip Localization from Stereo Images}
%
%
%


\author{Guijin~Wang,~\IEEEmembership{Senior Member,~IEEE,}
        Cairong~Zhang,
        Xinghao~Chen,
        Xiangyang~Ji,~\IEEEmembership{Member,~IEEE,}
        Jing-Hao~Xue,
        and~Hang~Wang
\thanks{G. Wang, C. Zhang and X. Chen are with the Department of Electronic Engineering, Tsinghua University, Beijing, 100084 China
(e-mail: wangguijin@tsinghua.edu.cn (G. Wang), zcr17@mails.tsinghua.edu.cn (C. Zhang), xinghao.chen@outlook.com (X. Chen)).}
\thanks{X. Ji is with the Department of Automation, Tsinghua University, Beijing, 100084 China (email: xyji@tsinghua.edu.cn).}
\thanks{J.-H. Xue is with the Department of Statistical Science, University College London, London, WC1E 6BT, UK (e-mail: jinghao.xue@ucl.ac.uk).}
\thanks{H. Wang is with Beijing Huajie IMI Technology Co., Ltd., Beijing, China (email: wanghang@hjimi.com).}
\thanks{C. Zhang and X. Chen are equally contributed to this work.}}

\maketitle

\begin{abstract}
In human-computer interaction, it is important to accurately estimate the hand pose especially fingertips. However, traditional approaches for fingertip localization mainly rely on depth images and thus suffer considerably from the noise and missing values.
Instead of depth images, stereo images can also provide 3D information of hands and promote 3D hand pose estimation. There are nevertheless limitations on the dataset size, global viewpoints, hand articulations and hand shapes in the publicly available stereo-based hand pose datasets. To mitigate these limitations and promote further research on hand pose estimation from stereo images, we propose a new large-scale binocular hand pose dataset called THU-Bi-Hand, offering a new perspective for fingertip localization. In the THU-Bi-Hand dataset, there are 447k pairs of stereo images of different hand shapes from 10 subjects with accurate 3D location annotations of the wrist and five fingertips. Captured with minimal restriction on the range of hand motion, the dataset covers large global viewpoint space and hand articulation space. To better present the performance of fingertip localization on THU-Bi-Hand, we propose a novel scheme termed Bi-stream Pose Guided Region Ensemble Network (Bi-Pose-REN).
It extracts more representative feature regions around joint points in the feature maps under the guidance of the previously estimated pose. The feature regions are integrated hierarchically according to the topology of hand joints to regress the refined hand pose. Bi-Pose-REN and several existing methods are evaluated on THU-Bi-Hand so that benchmarks are provided for further research. Experimental results show that our new method has achieved the best performance on THU-Bi-Hand.

\end{abstract}

\begin{IEEEkeywords}
Fingertip Localization, Hand Pose Estimation, Region Ensemble Network, Human-Computer Interaction, Hand Pose Dataset.
\end{IEEEkeywords}

%
\IEEEpeerreviewmaketitle

\section{Introduction}
\label{sec:intro}
\IEEEPARstart{A}{ccurate} pose estimation is one of the most essential technique in human-computer interaction (HCI), for example, virtual reality and augmented reality~\cite{erol2007vision,lee2014real,nolker2002visual,zhao2015learning}. Recently hand pose estimation from depth images has drawn a lot of attention from researchers~\cite{tompson2014real, supancic2015depth,tang2015opening, oberweger2015training, wan2016hand,wan2017crossing, yuan2018depth, moon2018v2v, ge20173d, zhang2018interactive}, thanks to the emergence of depth cameras~\cite{zhang2012microsoft, wang2013depth, shi2015high, keselman2017intel}. Fingertip localization is much more difficult than localizing other joints in a hand due to the large variation of viewpoints, high flexibility of fingertips, and poor depth quality around fingertips in depth images~\cite{tompson2014real}. Traditionally the stereo-based hand poses are estimated from stereo images by converting them into depth images using full stereo matching algorithms~\cite{rogister2012asynchronous}. However, depth values around fingertips may be inaccurate because large noise is introduced during conversion. As a result, fingertip localization is hindered due to error accumulation in the calculation of disparities and pose estimation from depth images.

Therefore, it is beneficial to estimate hand poses or fingertip locations from stereo images directly. Recently, several methods have been developed for fingertip localization from binocular images~\cite{chen2016accurate, wei2017two}, without converting them into depth images. However, the network architectures could not utilize the features of binocular images effectively. Furthermore, in the publicly available datasets of stereo-based hand poses~\cite{zhang2017hand, wei2017two}, there have been limitations on the dataset size, viewpoints, hand articulations and hand shapes. These limitations substantially limit the generalization ability of trained models.

In this paper, we propose a novel scheme termed Bi-stream Pose Guided Region Ensemble Network (Bi-Pose-REN), to localize the fingertips and the wrist from stereo images directly. Feature maps of left and right images are extracted using DenseNet~\cite{huang2017densely} in a two-stream style. Cropped around the location of joints in an initially estimated hand pose, grid feature regions of the two streams are then fused by concatenation, forwarded into fully connected (FC) layers, and integrated hierarchically according to the topology of hand joints. Under an iterative refinement framework, Bi-Pose-REN takes a previously predicted hand pose as input and improves the estimation in each iteration. Benefited from the ensemble learning of multiple branches and the more representative features for joint points, our proposed method has achieved state-of-the-art performance over existing methods on the ThuHand17~\cite{wei2017two} dataset.

To promote further research on hand pose estimation from stereo images, we build a large-scale binocular hand pose dataset, called THU-Bi-Hand, which contains about 447k pairs of stereo images from 10 different subjects with accurate annotations of six hand joint (five fingertips and the wrist) locations. In the dataset, 16 basic hand poses as well as transforming poses between pairs of basic poses were captured for each subject. The subjects were allowed to move their hands and fingers freely under the restriction that their hands appeared entirely in the valid imaging area. Captured from large diversity of hand shapes and hand poses, the new dataset sufficiently covers the natural hand pose space commonly used in HCI with little restriction on the range of hand motion including translation and rotation.

Our main contributions can be summarized as follows.

(1) We proposed a novel scheme, Bi-Pose-REN, for fingertip localization from stereo images directly, without conversion into depth images. Taking a previously estimated pose as input, Bi-Pose-REN extracts more representative feature regions of joint points and predicts more accurate results iteratively.

(2) To promote further research on fingertip localization from stereo images, we built a new large dataset THU-Bi-Hand~\footnote{Dataset available at https://sites.google.com/view/thubihand or http://image.ee.tsinghua.edu.cn/data/thubihand.}, which consists of about 447k stereo image pairs from 10 subjects with large variant hand poses and hand movements as well as accurate 3D location annotations of the fingertips and wrist.

(3) We provided several benchmarks on the THU-Bi-Hand dataset, offering a new perspective for fingertip localization. We evaluated several methods including Chen~et~al.~\cite{chen2016accurate}, TSBNet~\cite{wei2017two} and the proposed Bi-Pose-REN. Ablation studies of Bi-Pose-REN were also performed by analyzing the module effects.

The remainder of this paper is organized as follows. We review related work in Section~\ref{sec:related_work}. In Section~\ref{sec:bi-pose-ren}, we present details of our Bi-Pose-REN. In Section~\ref{sec:dataset}, we introduce the construction procedure and detailed information of the THU-Bi-Hand dataset. Experimental results and comparisons among different models are provided in Section~\ref{sec:experiment}. Section~\ref{sec:conclusion} concludes this paper and discusses the future work.

\section{Related Work}
\label{sec:related_work}
In this section, we will first review popular hand pose datasets. Next, we will discuss existing methods of hand pose estimation from stereo images. Finally, we will review different feature extraction methods used in hand pose estimation from depth images.

\subsection{Hand Pose Datasets}
\label{sec:review_dataset}
Publicly available hand pose datasets can be classified into two kinds: depth image based datasets~\cite{tompson2014real,tang2014latent,sun2015cascaded,yuan2017bighand2,wetzler2015rule} and stereo image based datasets~\cite{zhang2017hand,wei2017two}.

The NYU dataset~\cite{tompson2014real} contains over 72k RGB-D images from one subject in training set and 8k images from 2 different subjects (one of them is the subject in training set) for testing with 36 annotated joints. The depth maps were collected from Microsoft Kinect camera~\cite{zhang2012microsoft} with missing values along occluded boundaries and noisy outlines~\cite{oberweger2015hands}.

The ICVL dataset~\cite{tang2014latent} has 300k images with different rotations from 10 subjects with 26 gestures for training and 1.6k images for testing. The depth images were captured by Intel RealSense~\cite{keselman2017intel} with locations of 16 joints annotated. The depth maps have a high quality with little missing values and sharp outlines with little noise. But lots of samples were annotated incorrectly in both training and test sets (about 36\% of the poses from the test set were annotated with an error of at least 10mm)~\cite{oberweger2015hands}.

The MSRA dataset~\cite{sun2015cascaded} contains 76.5k depth images collected with Intel Creative Interactive Camera. Totally 21 joints were annotated. There are 9 subjects with 17 gestures for each subject. The variation of hand poses is limited in this dataset.

The BigHand2.2M~\cite{yuan2017bighand2} contains 2.2 million depth maps with accurately annotated 21 joint locations. It has large diversity of global viewpoint, hand articulation and orientation due to its minimal restriction on the range of motions. There are 10 hand shapes in training set and an additional shape for testing.

The HandNet dataset~\cite{wetzler2015rule} was created from 10 participants, containing more than 210k depth images captured by Intel RealSense camera, with annotations of the center of the hand and the five fingertips.

There are also several attempts for stereo hand pose datasets. In~\cite{zhang2017hand}, a stereo hand pose dataset was established containing 18k stereo image pairs with annotations of palm and finger joints. The images were captured by a Point Grey Bumblebee2 stereo camera, divided into 12 different sequences. There is only one subject in this dataset. It is too small and contains only one hand shape, which limits the generalization ability of trained models.

In the ThuHand17 dataset~\cite{wei2017two}, the training set contains about 117k pairs of binocular images of 8 subjects, with 16 basic hand poses and a lot of transitional poses between adjacent basic poses captured by Leap Motion. Seven subjects performed all the 16 basic poses and the transitional poses, while the remaining one mainly performed several kinds of basic poses. The test set contains 10k pairs of binocular images (different from training samples) of 2 subjects. However, ThuHand17 is still not large enough, and the 2 subjects of its test set are included in the 8 subjects of its training set.

Currently, publicly available datasets of stereo-image-based hand poses are insufficient for fingertip localization. In order to promote research of fingertip localization from stereo images, we built a new large-scale binocular hand pose dataset called THU-Bi-Hand. Totally about 447k stereo images from 10 different subjects were collected. All samples of seven subjects and half of the samples of another two subjects construct the training set, while the remaining samples (including half of the samples of two subjects and all the samples of one subject) form the test set. There are about 357k and 90k samples in the training set and test set respectively. THU-Bi-Hand is the largest binocular hand pose dataset with large variety of hand poses, hand movements and hand shapes.

\subsection{Stereo-based Hand Pose Estimation}
\label{sec:review_pose}
Recent approaches to hand pose estimation from stereo images can be categorized into two categories: indirect methods~\cite{zhang2017hand,basaru2017hand} and direct methods~\cite{romero2008dynamic,chen2016accurate,panteleris2017back,wei2017two}. Indirect methods first compute depth maps from stereo images, and then estimate hand poses from depth images. Direct methods estimate hand poses directly from stereo images, without converting them into depth maps.

The indirect method of Zhang~et~al.~\cite{zhang2017hand} incorporates on-line training based skin color detector and constrained stereo matching to compute depth maps from stereo images and conduct hand segmentation. Then, depth-based hand pose tracking algorithms~\cite{qian2014realtime, oikonomidis2011efficient} are utilized to estimate hand poses in stereo image sequences. However, it still cannot get rid of poor depth quality around fingertips, which will cause difficulties in fingertip localization. Furthermore, the error in depth map calculation from stereo images hinders the performance of depth-based hand pose estimation.
In~\cite{basaru2017hand}, depth proposals and hand poses are jointly optimized using Markov-chain Monte Carlo (MCMC) sampling and two CNNs. The first CNN evaluates the consistency between the proposed depth images and the observed stereo images, while the second CNN estimates hand poses from the proposed depth images.
However, it also suffers from poor depth quality around fingertips as~\cite{zhang2017hand}. Besides, it consumes much time with a lot of depth proposals. A frame of stereo images under 200 MCMC proposals takes 360 seconds during prediction.

Direct methods can avoid the influence of noise introduced during depth map conversion.
In~\cite{panteleris2017back}, a generative hand model based framework is proposed to optimize the hand pose that maximizes the color consistency of the two views of the hand, avoiding the explicit computation of disparity maps of relatively uniformly colored hands. However, an explicit definition of the hand model is required for model-driven methods. It is also sensitive to the initialization of hand poses and suffers from tracking failures.
In~\cite{chen2016accurate}, hand mask images extracted from binocular images are exploited to predict the 3D positions of fingertips and the palm center using a deep convolution neural network (CNN) without explicitly computing depth maps. But the network architecture cannot utilize the features of binocular images effectively. Besides, lots of information is lost with only mask images as input.
In~\cite{wei2017two}, both original images and hand mask images are used as the input. Low-level feature maps of both left and right images are extracted by convolutional layers with the same structure and shared parameters, while high level features of the two streams are extracted by CNN layers with the same structure but different parameters separately. The features from two streams are fused by FC layers to estimate the pixel coordinates of the joints. However, the whole feature maps are used to regress joint positions, which is not optimal for each joint. Localized features can be used for better estimation.

\subsection{Feature Extraction And Region Ensemble}
\label{sec:review_feature}
CNN based architectures are proved to be very powerful in many computer vision tasks due to their strong ability of image based feature extraction~\cite{krizhevsky2012imagenet,simonyan2014very,szegedy2015going,he2016deep,huang2017densely}. In~\cite{he2016deep}, residual representations and shortcut connections are incorporated into CNNs to address the problem of accuracy degradation when networks go deeper.
By inserting identity shortcut connections between convolutional layers, the network is forced to learn residual mapping, which is beneficial for improving performance in deeper networks. Residual connections also ease the optimization by providing faster convergence for relatively shallow networks.

In~\cite{huang2017densely}, the benefits of connections between layers are further utilized to formulate the densely connected convolution networks (DenseNets). DenseNets have several dense blocks connected by transition layers. Inside each dense block, every layer is connected to all other layers.
For each layer in a dense block, the feature maps of all preceding layers are concatenated and used as inputs, and its own feature maps are used as inputs into all subsequent layers after being concatenated with feature maps of other layers. The dense connections and feature map concatenations can alleviate the vanishing-gradient problem, strengthen feature propagation and encourage feature reuse. But DenseNets are highly memory consuming because of fast feature maps growing.

In~\cite{guo2017region,wang2018region,chen2017pose}, CNNs with residual connections are used for feature extraction of hand pose estimation and achieve promising results. Furthermore, in~\cite{guo2017region,wang2018region}, region ensemble network (REN) has been proposed to exploit good practices and improve performance for hand pose estimation from depth images. With feature maps of the last convolutional layer, REN divides them into several grid regions. Region ensemble strategy is utilized by concatenating the FC layer outputs of different regions, which can represent multiple views of input images. Benefited from the multi-view strategy in both training and testing as well as ensemble learning of multiple branches, REN achieves a great improvement in depth-based hand pose estimation.

In~\cite{chen2017pose}, the region ensemble method is further utilized to generate the pose guided structured region ensemble network (Pose-REN) to boost the performance of hand pose estimation from depth images. Pose-REN is an iterative refinement procedure, taking a previously estimated pose as input and predicting a more accurate result in each iteration. It crops spatial regions from the feature maps around each joint of the previously predicted hand pose. The cropped feature regions are integrated hierarchically according to the topology of hand joints by FC layers and produce a refined hand pose, which is used as a guidance for feature cropping in the next iteration.

\section{Bi-Pose-REN}
\label{sec:bi-pose-ren}
In this section, we describe the proposed Bi-Pose-REN, which estimates the positions of fingertips and wrist from stereo images directly without conversion into depth images.

\begin{figure*}[htb]
  \centering
  \includegraphics[width=1.0\textwidth]{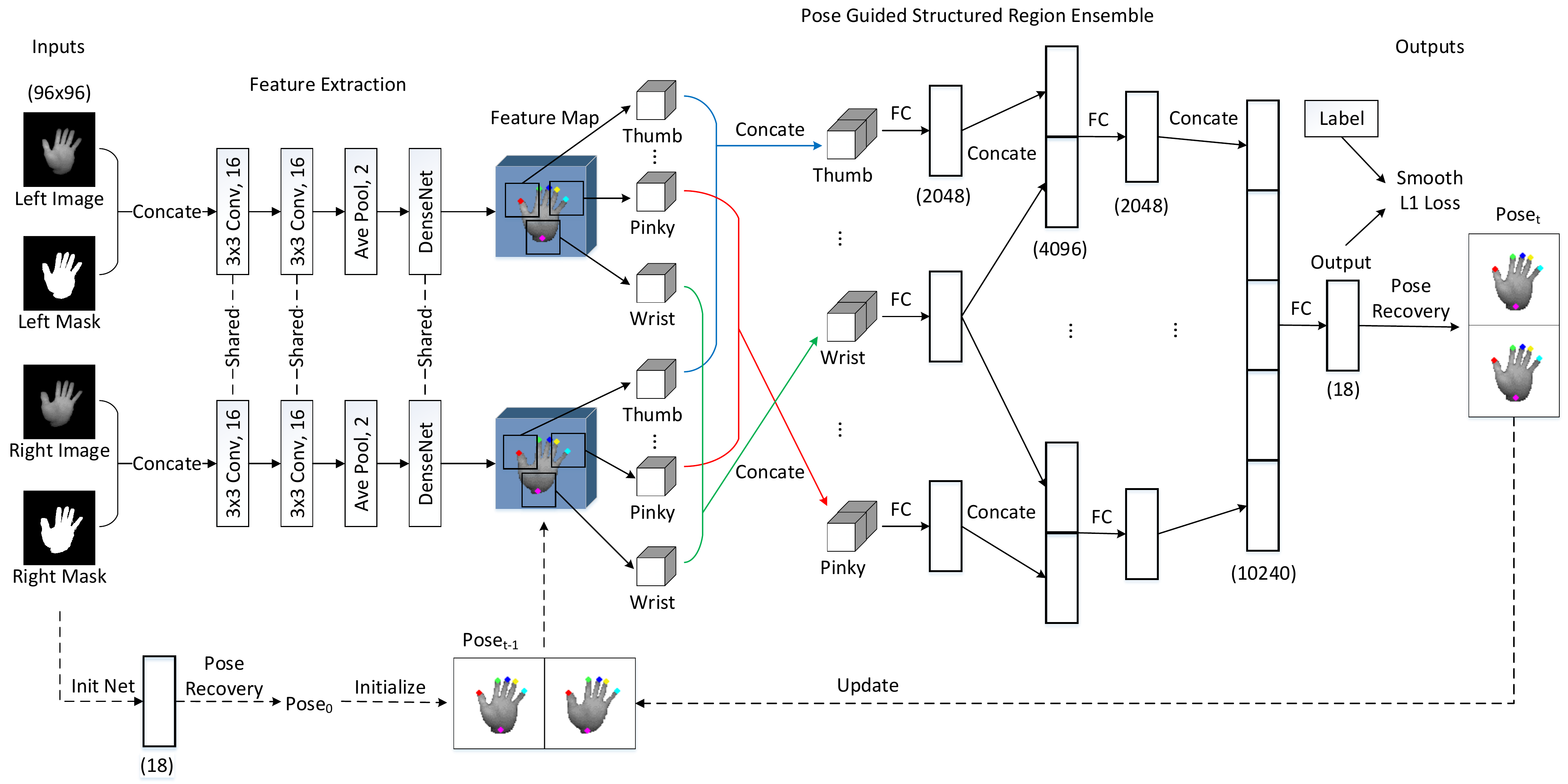}\\
  \caption{The framework of our Bi-stream Pose Guided Region Ensemble Network (Bi-Pose-REN). ``$k \times k \ Conv, c$'' represents a convolutional layer with a $k \times k$ size kernel and $c$ channels ($stride = 1, pad = 1$). ``$Ave \ Pool, 2$'' denotes an average pooling layer with kernel size $2 \times 2 $ ($stride = 2, no \ padding$). ``$FC$'' represents a fully connected layer. The numbers inside parentheses denote the sizes of images or vectors.}\label{fig:framework}
\end{figure*}

As shown in Fig.~\ref{fig:framework}, Bi-Pose-REN uses cropped masks and stereo images as input. Wei~et~al. pointed out that masks are robust to variation of illumination and skin colors, while original images contain more information of the hand~\cite{wei2017two}. So using both cropped masks and stereo images as input can enhance the robustness and the precision of our model.

First, left image and left mask are concatenated, as well as right image and right mask. Left inputs and right inputs are processed in two different streams with the same structure and shared parameters. Then, in the feature extraction module, DenseNet structured layers are employed to extract feature maps of left input and right input (both in size $96 \times 96 \times 2$) in two streams separately. In contrast to~\cite{huang2017densely}, in Bi-Pose-REN, before being passed to dense blocks, the input images are first forwarded to convolutional layers and an average pooling layer, which eases the consumption of GPU memory. Furthermore, considering that batch normalization (BN)~\cite{ioffe2015batch} helps little in our tasks in practice, we remove BN layers used in standard DenseNet in order to improve the performance, accelerate training and inference procedures, as well as reduce memory consumption.

Later on, inspired by Pose-REN~\cite{chen2017pose}, we utilize the pose-guided region ensemble method to localize fingertips from stereo images. Feature regions are extracted from the feature maps around previously estimated joint locations in each stream, and every two corresponding regions of the left and right feature maps are fused by concatenation. Then the concatenated regions are integrated by FC layers hierarchically according to the topology of hand joints. Finally, transforming the 18-dimensional outputs into hand poses through another FC layer and an additional pose recovery layer, Bi-Pose-REN predicts hand poses end-to-end. We use the rectified linear unit (ReLU)~\cite{glorot2011deep} as the activation function.

Bi-Pose-REN refines joint locations iteratively given an initial hand pose. The initialization network (Init Net) is designed to predict the initial hand pose in Bi-Pose-REN. To explore the robustness of Bi-Pose-REN, different Init Nets are studied during inference.

Bi-Pose-REN differs from our previous Pose-REN~\cite{chen2017pose} (for depth images) in three aspects. First, Bi-Pose-REN employs DenseNet~\cite{huang2017densely} in feature extraction instead of CNN with residual connections. Second, Bi-Pose-REN is a two-stream network, which extracts feature maps of left inputs and right inputs in two streams. Furthermore, after grid regions extraction from two-stream feature maps, corresponding regions are concatenated to fuse the information in different streams.

\subsection{Preprocessing}
\label{sec:preprocess}
We give a brief introduction to our preprocessing procedure. Since the stereo images in the THU-Bi-Hand dataset were captured with the infrared imaging device, Leap Motion~\cite{leapmotion}, the hand regions can be extracted by a thresholding method.

Two different thresholds $th_1$ and $th_2$ were set considering the trade-off between preserving the completeness of the entire hand and removing the noise in the background. The larger one ($th_1$) is first used to obtain rough binary mask images and only the connected components with the maximum area are preserved to calculate the centroids of hand regions. Stereo images are cropped around the centroids. Then we use the same method but with $th_2$ to get delicate binary mask images from cropped stereo images. For convenience, the centroids of the left and right images are denoted as $(c_{xl}, c_{yl})$ and $(c_{xr}, c_{yr})$ respectively ($c_{yr}$ and $c_{yl}$ are set to equal).

Cropped stereo images and cropped mask images are resized into a predefined size $w_p \times h_p$ before being fed into the networks. In cropped stereo images, the pixels outside the hand masks are set to zero, while other pixels remain unchanged.

\subsection{Inputs}
\label{sec:inputs}
The input of Bi-Pose-REN includes four images: the stereo images and the mask images (both left and right) after preprocessing. Besides, in order to increase the model robustness of different sizes of hand shapes, we use multi-scale training. Three sizes of images are cropped from the original stereo images and binary mask images: $240 \times 240$, $220 \times 220$, $200 \times 200$. So the training set is enlarged three times. While testing, the size of $200 \times 200$ is used to crop the images. The cropped left image and mask are concatenated into a two-channel input, the corresponding cropped right image and right mask are concatenated as well.

Data augmentation: To further increase the robustness of Bi-Pose-REN, we perform data augmentation in two ways: translation and zooming. The images are translated for random $(m, n)$ pixels along both horizontal and vertical directions, and zoomed by a factor $s$ during training ($m$, $n$ and $s$ are sampled from uniformed distributions with predefined borders). The ground truths of the six joints are transformed correspondingly.

\subsection{Bi-stream DenseNet Structure for Feature Extraction}
\label{sec:densenet}
We use two streams with the same structure and shared parameters to extract feature maps of left and right images separately. Bi-Pose-REN incorporates the DenseNet structure into feature extraction. But unlike~\cite{huang2017densely}, in Bi-Pose-REN, two convolutional layers and an average pooling layer are used in front of the first dense block. The BN layers used in the standard DenseNet are also removed in Bi-Pose-REN. Besides, we set a small growth rate and fewer layers in a dense block than~\cite{huang2017densely}.

Specifically, in our Bi-Pose-REN, the growth rate of a dense block is 24. There are 3 dense blocks, each containing 2 convolutional layers with kernel size $3 \times 3$, stride 1 and zero-padded to keep the size of feature maps fixed. The first convolutional layer in the first dense block has 16 output channels. Between two contiguous dense blocks, a $1 \times 1$ convolution and a $2 \times 2$ average pooling layer with stride 2 are set as the transition layers. Considering the first two convolutional layers before dense blocks, there are totally 10 convolutional layers in each stream in our Bi-Pose-REN.

\subsection{Pose Guided Structured Region Ensemble Network}
\label{sec:pose-ren}
\begin{figure}[htb]
  \centering
  \includegraphics[width=1.0\columnwidth]{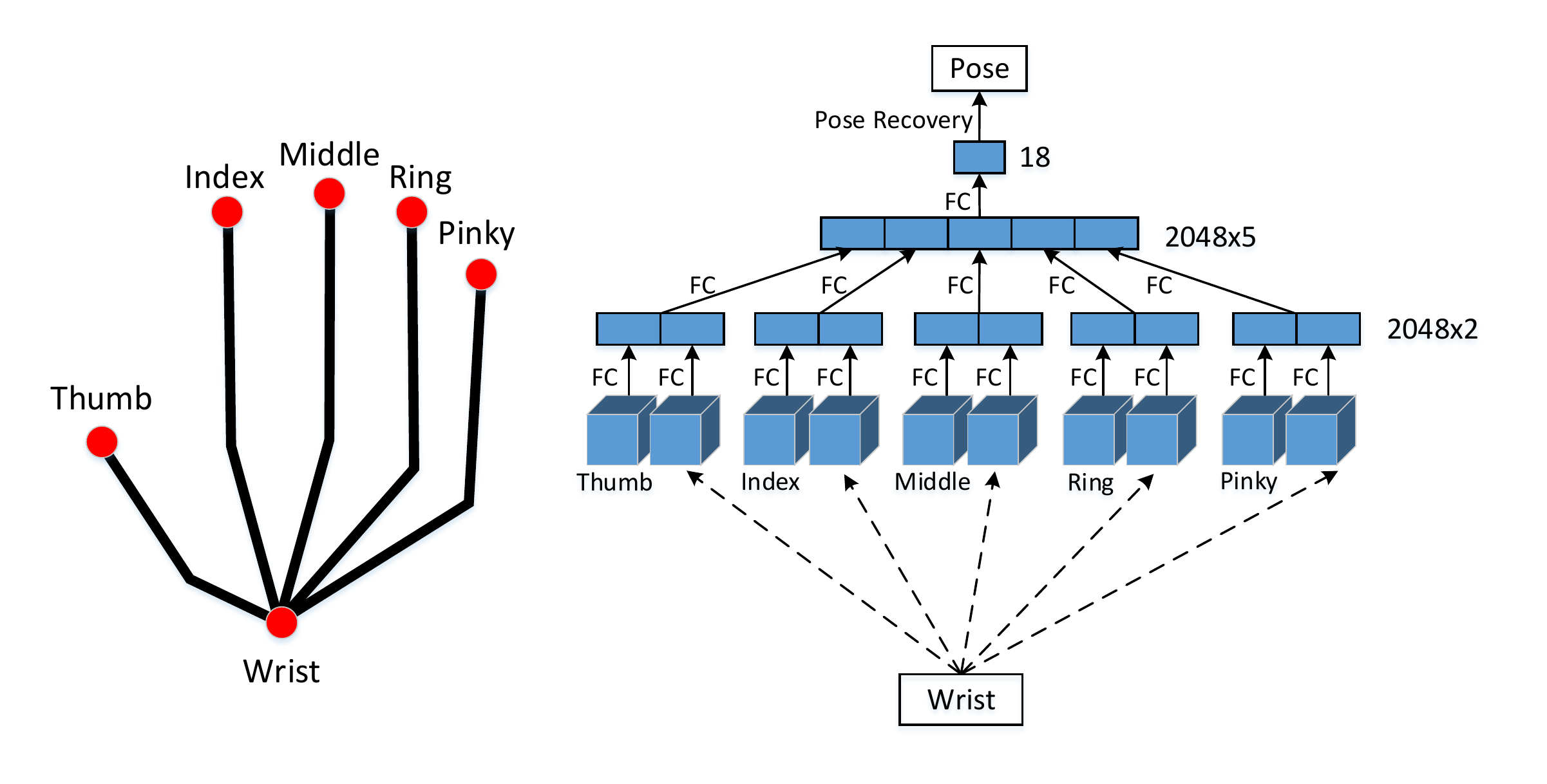}\\
  \caption{The hand topology (left) and the architecture of the proposed region ensemble method (right). Each fingertip is fused with the wrist first. Afterwards, FC (fully connected layer) outputs of different fingers are fused to regress the final hand pose. }
  \label{fig:topology}
\end{figure}

Inspired by Pose-REN~\cite{chen2017pose}, which promoted the accuracy in hand pose estimation from depth images, we further extend the idea of pose guided region ensemble for fingertip localization from stereo images. We estimate an initial hand pose using the initial network (Init Net). The initial hand pose is used as the guidance in the first iteration in Bi-Pose-REN. After feature extraction with two stream DenseNet structured CNN in Section~\ref{sec:densenet}, the feature maps are cropped into several regions around each joint location according to the initial hand pose in both left and right streams separately. For each cropping center, the regions of the left and right feature maps are concatenated along the channel dimension to fuse the information of two streams, generating six concatenated regions. Then the concatenated regions of the six joints are integrated hierarchically according to their topology, with the wrist and each fingertip concatenated first followed by the five finger parts concatenated later (See Fig.~\ref{fig:topology} (right)). FC layers are used to fuse the information of concatenated feature regions and predict the refined hand pose, which is used later to guide the feature maps cropping procedure in the next iteration. In this manner, the estimated hand pose can be refined iteratively under the pose guided framework.

Denote the pixel coordinates of hand joints in the left and right images as $(u_l, v_l)$ and $(u_r, v_r)$ respectively, where $v_l$ equals to $v_r$. These original coordinates are not consistent with the inputs of Bi-Pose-REN. Therefore, in order to reduce the difficulties of mapping cropped stereo images to some variables corresponding to the positions of joints, we use the easier mapping form (Eq.\ref{eq:label}):
\begin{equation}\label{eq:label}
  label =
  \begin{pmatrix}
  \frac{(u_l - c_{xl}) + (u_r - c_{xr})}{w} \\
  \frac{2((u_l - c_{xl}) - (u_r - c_{xr}))}{w} \\
  \frac{(v_l - c_{yl}) + (v_r - c_{yr})}{h}
  \end{pmatrix},
\end{equation}
where $(c_{xl}, c_{yl})$ and $(c_{xr}, c_{yr})$ are the centroids of the segmented hand region in the left and right images respectively, and $w$ and $h$ are the width and height of the cropped images (before resized). $(u_l + u_r)$ and $(u_l - u_r)$ are the elements corresponding to the horizontal coordinates and disparities of joint positions. $(v_l + v_r)$ is corresponding to the vertical coordinates. The coordinates of each joint in cropped images (with centroids subtracted) are normalized by $w$ and $h$.

Bi-Pose-REN minimizes the smooth L1 loss~\cite{girshick2015fast} between the 18-dimensional output of the last FC layer and the transformed label (Eq.~\ref{eq:label} of six joints concatenated), with the threshold as $th=0.01$ (Eq.~\ref{eq:loss}):
\begin{equation}\label{eq:loss}
  smooth_{L1}(x) =
  \begin{cases}
    0.5|x|^2/th, &\text{ if }|x|<th\\
    |x|-0.5th, &\text{ otherwise }
  \end{cases}.
\end{equation}
Afterwards, a pose recovery layer is added to transform the outputs of the last FC layer into hand poses (in the form of $(u_l, v_l)$ and $(u_r, v_r)$, with $v_l$ equals to $v_r$).

\section{Dataset: THU-Bi-Hand}
\label{sec:dataset}
In this section we introduce our method to build the THU-Bi-Hand dataset using Leap Motion~\cite{leapmotion} to capture binocular images of hands with a resolution of $640\times480$, and the TrakSTAR tracking system with 6D magnetic sensors~\cite{trakstar} to obtain accurate annotations of the locations of the wrist and five fingertips. We also present detailed information about the THU-Bi-Hand dataset, including the size of the dataset, etc.

\subsection{Dataset Construction}
\label{sec:construct}
Our hand model has 6 joints: the wrist and five fingertips (See Fig.~\ref{fig:topology} (left)). To avoid the ambiguous definition of the wrist, we consistently define the location of the wrist as the intersection of two lines: the first one is the bone connecting the root of the middle finger and the center of the palm; the second one is perpendicular to the first line and passes through the root of the thumb.

\begin{figure}[htb]
  \centering
  \includegraphics[width=0.6\columnwidth]{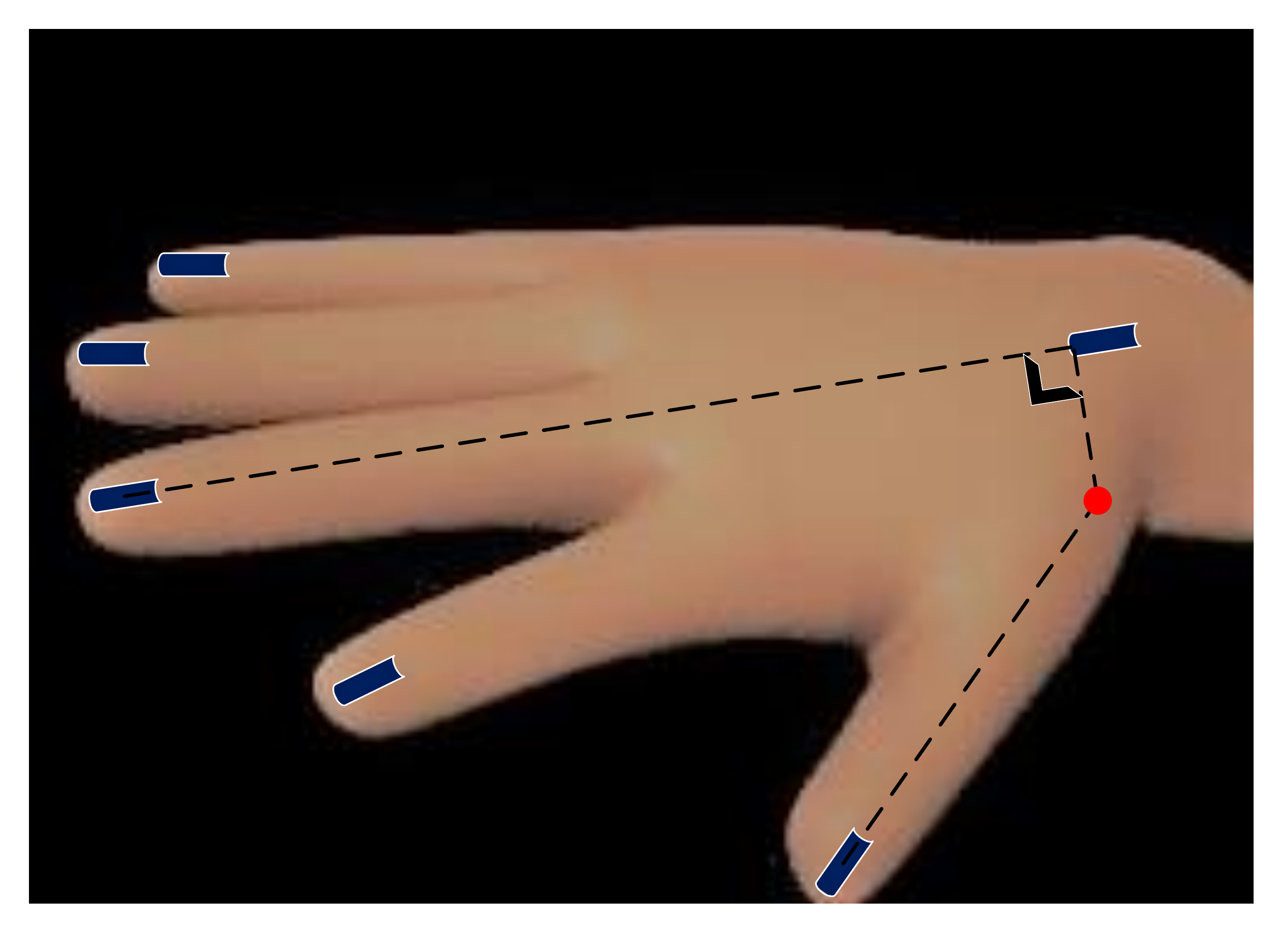}\\
  \caption{Annotation setting. The sensors were attached to the back of the hand, on the locations of five fingertips and the wrist.}
  \label{fig:setting}
\end{figure}

\begin{figure}[htb]
  \centering
  \includegraphics[width=1.0\columnwidth]{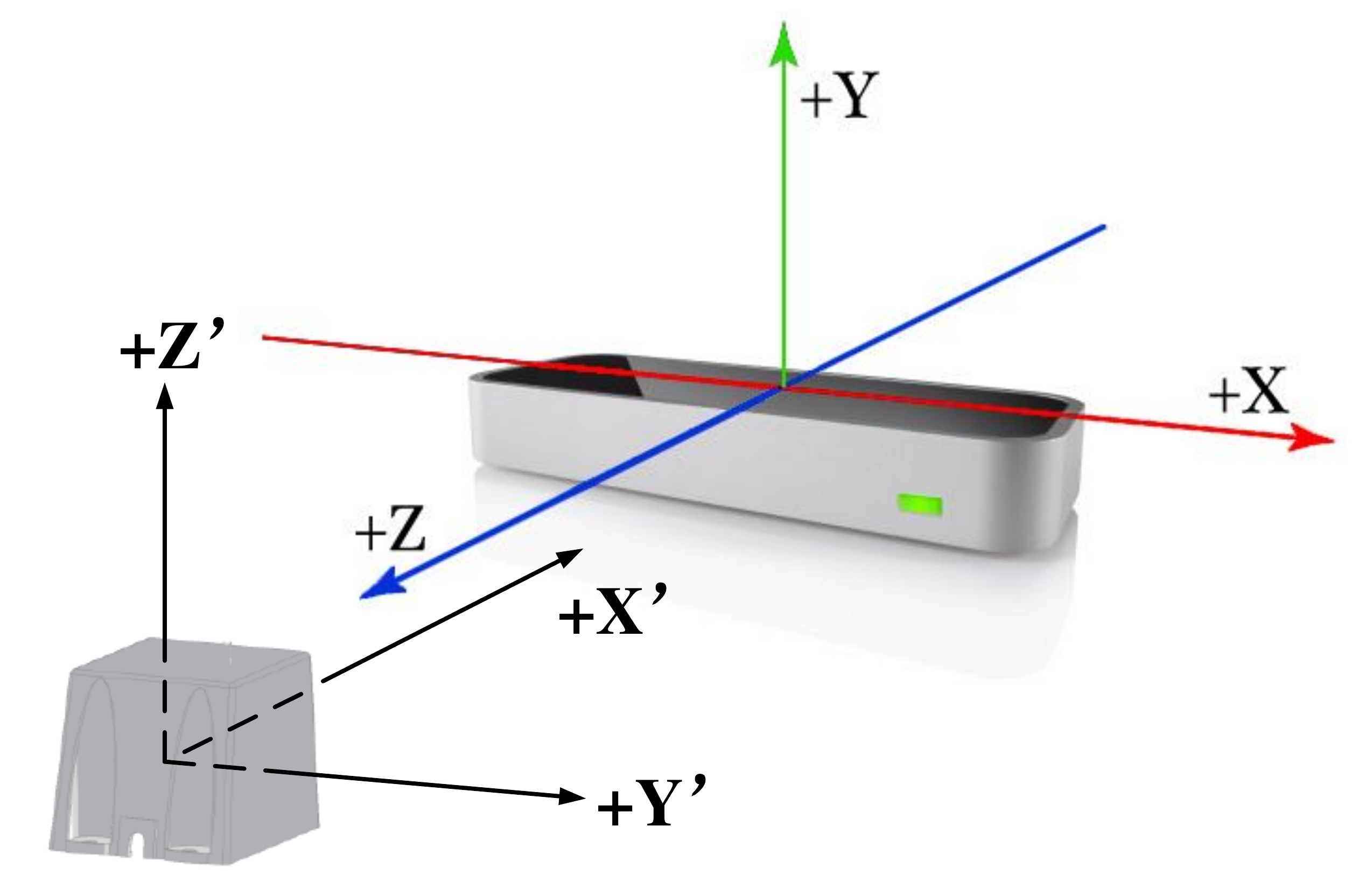}\\
  \caption{The coordinate systems of TrakSTAR~\cite{trakstar} and Leap Motion~\cite{leapmotion}. The devices were placed parallel to each other.}
  \label{fig:coordinate}
\end{figure}

To build and annotate the THU-Bi-Hand dataset, we used Leap Motion and the TrakSTAR system with six 6D magnetic sensors. Leap Motion is a kind of infrared imaging binocular camera, so we can perform segmentation easily using thresholds of grayscale. The TrakSTAR system can track the attached magnetic sensors by capturing their orientations and locations under the precision of 1.4mm. As shown in Fig.~\ref{fig:setting}, we attached the six sensors to the back of the hand, located on the defined wrist and five fingertips. Note that the locations of the magnetic sensors are a bit different from the real locations of the six joints since the joints are located inside the hand or fingers, not on the surface of the back. The coordinates of the six sensors in the coordinate system of the TrakSTAR were acquired, while the coordinates we really needed were the ones in the coordinate system of the Leap Motion (Fig.~\ref{fig:coordinate}). To simplify the problem, the Leap Motion and the Transmitter of the TrakSTAR system were placed parallel to each other. Then we measured the position of the coordinate origin of the Leap Motion using one magnetic sensor, afterwards we can do calibration easily since there is no explicit rotation (Eq.~\ref{eq:calibration}):
\begin{equation}\label{eq:calibration}
  \begin{aligned}
  & x = y' - y'_0 \\
  & y = z' - z'_0 \\
  & z =-x' + x'_0
  \end{aligned},
\end{equation}
where $(x', y', z')$ and $(x, y, z)$ are coordinates in the TrakSTAR coordinate system and the Leap Motion system respectively, and $(x'_0, y'_0, z'_0)$ is the coordinates of the origin of the Leap Motion in the TrakSTAR coordinate system.

For better segmentation, considering that Leap Motion is based on infrared imaging, volunteers were asked to wear black wrist straps. The TrakSTAR system consists of two electronic magnetic units, which are synchronized using Multi-Unit Sync Connections~\cite{trakstar}. Each magnetic unit can attach at most four magnetic sensors, which are 2mm wide, with a 1.2mm wide and 3.3m long cable. The cables were attached to the hand using small and short tubes so they can move freely through the tubes and the hand movements were less affected. The images and annotations were synchronized as in~\cite{yuan2017bighand2}.

\begin{figure}[htb]
  \centering
  \includegraphics[width=0.98\linewidth]{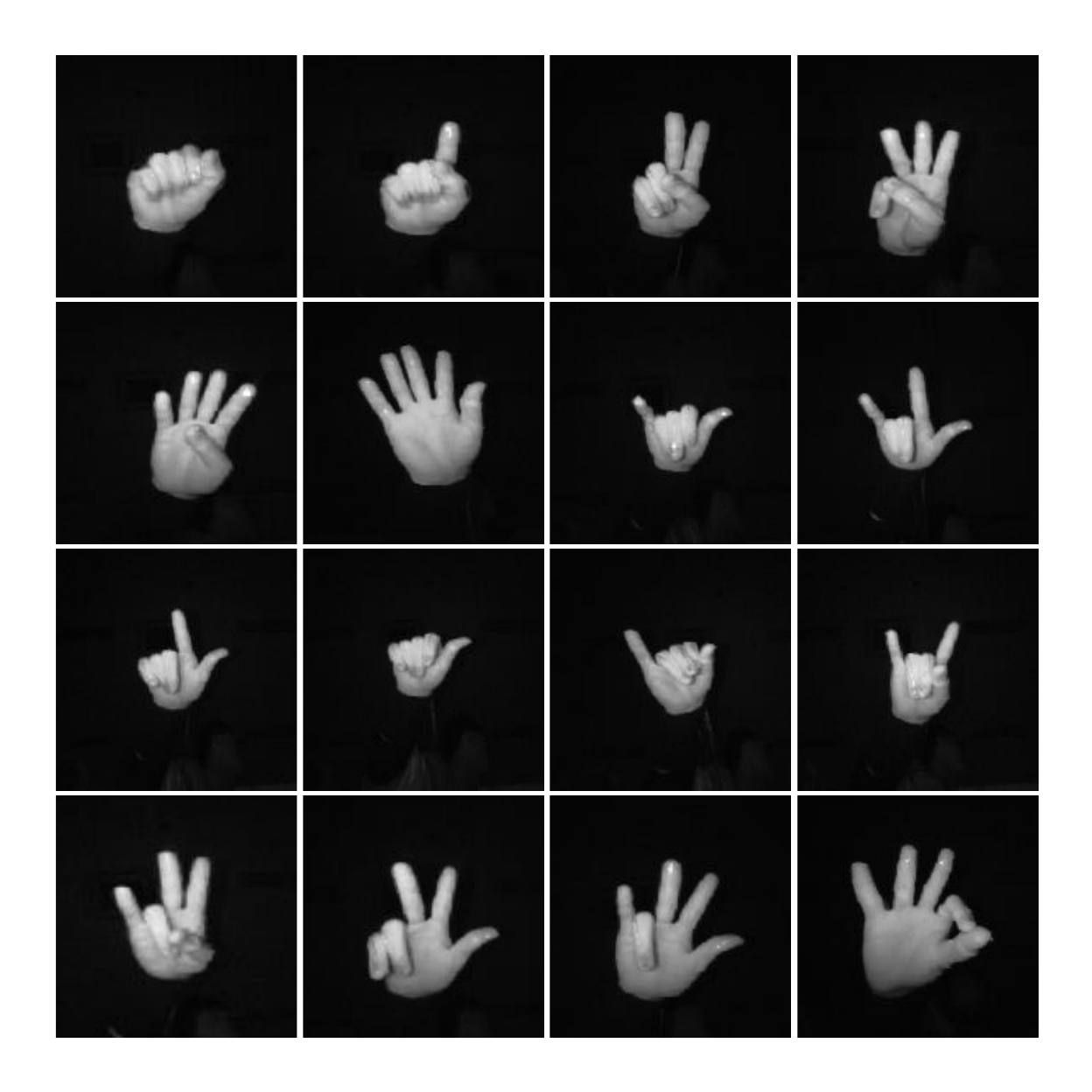}
  \caption{The 16 basic hand poses in the THU-Bi-Hand dataset.}
  \label{fig:basic_poses}
\end{figure}

There are 16 basic hand poses in the THU-Bi-Hand dataset, see Fig.~\ref{fig:basic_poses}. Each subject was asked to perform the basic hand poses in the order as shown in Fig.~\ref{fig:basic_poses}, with both translation and rotation of the hand. The transforming poses between each pair of the adjacent basic poses were also captured, and the subject was asked to perform the transforming procedure several times before he performing the next basic pose. Similarly, translation and rotation of the hand were also involved during the transforming procedure. After all basic poses were performed, another similar procedure was carried out but in the reverse order of the 16 basic poses. Afterwards, the subject was asked to do transformations between several pairs of basic poses that were not adjacent, for example, changing between the second basic pose and the fourth basic pose, the second pose and the fifth pose, etc. The pairs of poses were chosen randomly by the subject himself. Besides, during the sampling procedure, the subject could move his fingers freely, like swinging and clicking. Translations along all three directions were allowed as long as the hand did not disappear from the valid imaging area. Rotations within 90 degrees around all the three axis were allowed, while the initial hand was in the plane parallel to the imaging plane of the Leap Motion and extended perpendicular to the baseline connecting the left camera and the right camera of the Leap Motion.

\subsection{Glimpse of the THU-Bi-Hand Dataset}
\label{sec:glimpse}
Similar to~\cite{wei2017two}, we defined 16 basic hand poses (See Fig.~\ref{fig:basic_poses}). In order to better cover the articulation space, hand pose changing between two basic poses were also captured. As mentioned in Section~\ref{sec:construct}, the THU-Bi-Hand dataset consists of two parts: (1) Basic poses: 16 kinds, almost fully cover all hand poses used in HCI. For each subject, about 1000 or more frames were captured for each basic pose. (2) Transforming poses: contains hand poses while subjects transformed their hand pose from one basic pose to another, including transforming poses between each pair of adjacent basic poses, about 800 frames for each pair, and more than 10000 frames of transforming poses between those pairs of basic poses which are not adjacent.

THU-Bi-Hand contains samples of ten subjects with different hand shapes, while the numbers of subjects in the datasets of~\cite{chen2016accurate} and~\cite{wei2017two} are only one and eight respectively. Totally about 447k frames of the left and right images with locations of six joints annotated were captured, which is almost 3 times larger than the ThuHand17 dataset~\cite{wei2017two}. In the experiments, we use a subset containing all samples of seven subjects and half of the samples of another two subjects for training, while the remaining samples (including half of the samples of two subjects and all the samples of the remaining one subject) are used for testing. The training set and the test set contain about 357k and 90k frames respectively. The subject for testing has a different hand shape from the subjects for training, so we can test the generalization ability of different methods. We must point out that in~\cite{wei2017two}, some hand poses in the test set have a larger range of rotation and translation than the train set, but the hand shapes in their test set also appear in the train set, which is unfavorable while testing models' generalization ability.

\begin{figure*}
  \centering
  \includegraphics[width=0.97\textwidth]{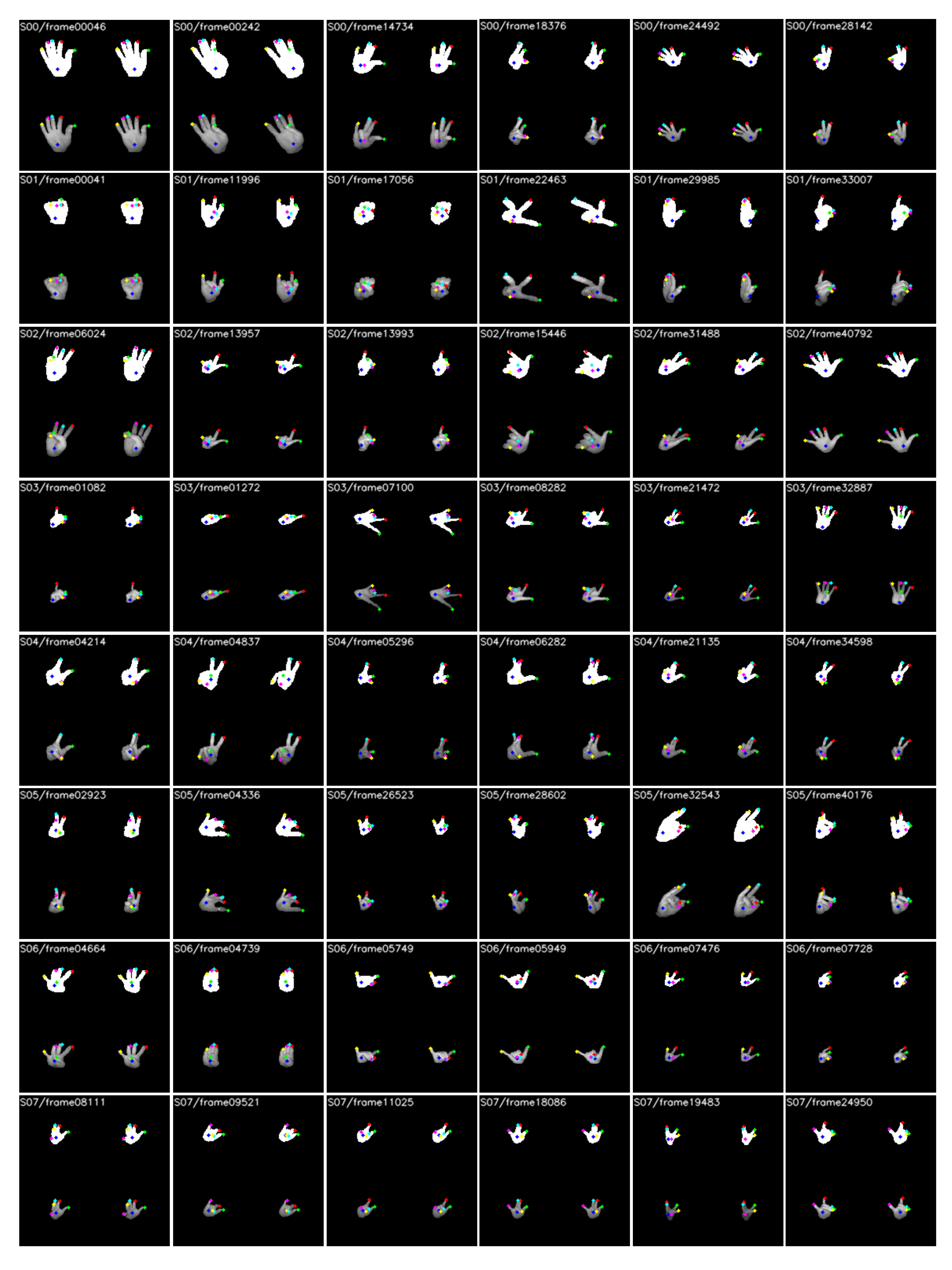}
  \caption{Some examples from THU-Bi-Hand dataset. Cropped masks and stereo images (both left and right) with annotations after preprocessing are shown.}
  \label{fig:examples}
\end{figure*}

The THU-Bi-Hand dataset has a large variety of hand poses, large range of hand movements: rotation and translation, large diversity of hand shapes and large size of data samples. THU-Bi-Hand can be very beneficial for promoting research in hand pose estimation, especially fingertip detection from binocular images. Fig.~\ref{fig:examples} shows some example images (after preprocessing) and corresponding masks in THU-Bi-Hand.

\subsection{Dataset Analysis}
\label{sec:dataset_analysis}
We compare the THU-Bi-Hand dataset and the hand pose benchmark in~\cite{zhang2017hand} by visualizing the global viewpoints, hand articulation and hand shapes of both of the data using t-SNE visualization~\cite{maaten2008visualizing,van2014accelerating} and PCA projection.

\subsubsection{Hand Viewpoint Space}
During data sampling, the subjects were asked to explore the global viewpoint space as large as possible by moving their hands in all the directions, under the constraint that the hand region captured by the camera is neither too small nor too big. They were also asked to rotate their hands randomly, as long as the rotation angles were smaller than 90 degrees from the natural state facing the camera. As shown in Fig.~\ref{fig:tsne} (left), the THU-Bi-Hand dataset covers much larger hand viewpoint space than the stereo dataset in~\cite{zhang2017hand}.

\subsubsection{Hand Articulation Space}
\begin{figure*}[htb]
  \begin{minipage}[b]{0.33\textwidth}
    \centering
    \centerline{\includegraphics[width=1.0\textwidth]{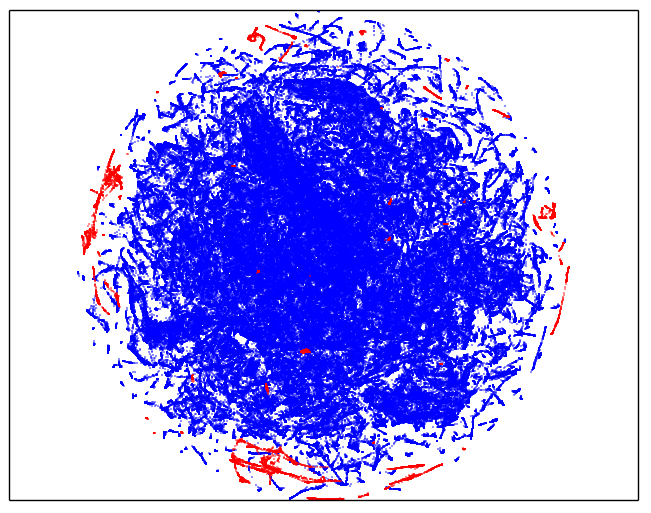}}
  \end{minipage}
  \begin{minipage}[b]{0.33\textwidth}
    \centering
    \centerline{\includegraphics[width=1.0\textwidth]{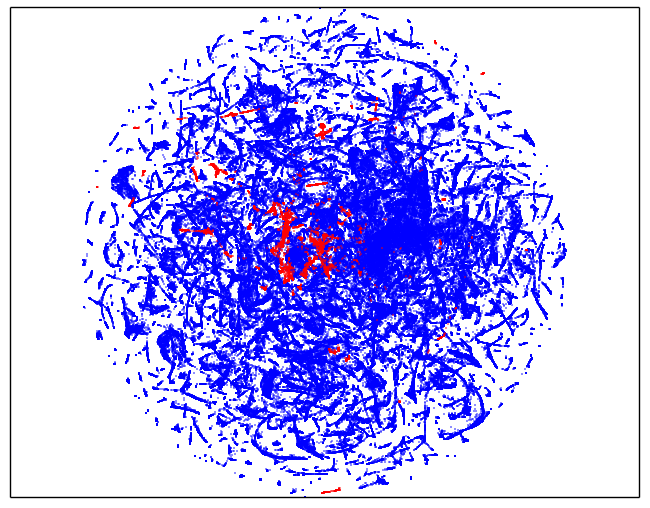}}
  \end{minipage}
  \begin{minipage}[b]{0.33\textwidth}
    \centering
    \centerline{\includegraphics[width=1.0\textwidth]{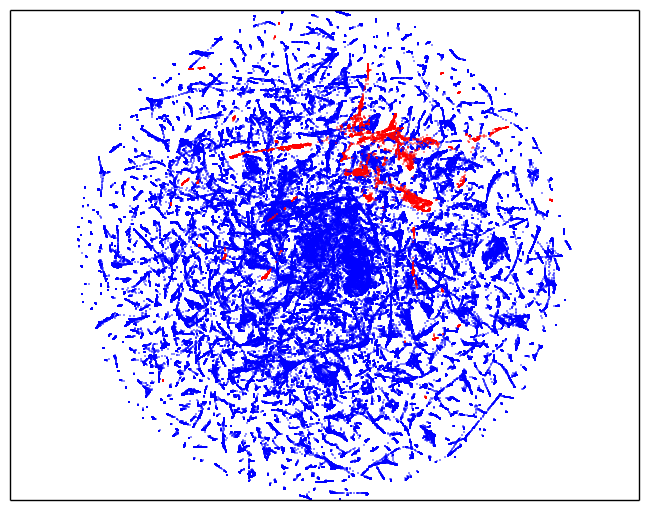}}
  \end{minipage}
  \caption{2D t-SNE visualization of the hand pose space. THU-Bi-Hand and Zhang~et~al.~\cite{zhang2017hand} datasets are represented in blue and red respectively. Left: global viewpoint space. Middle: hand articulation space. Right: combination of global viewpoint and hand articulation coverage. THU-Bi-Hand covers much larger hand pose space compared with Zhang~et~al.~\cite{zhang2017hand}.}
  \label{fig:tsne}
\end{figure*}

In THU-Bi-Hand dataset, all subjects were asked to perform 16 basic hand poses and abundant transforming poses between pairs of basic poses. Compared with~\cite{zhang2017hand}, the THU-Bi-Hand dataset explores much larger hand articulation space (See Fig.~\ref{fig:tsne} (middle and right)).

\subsubsection{Hand Shape Space}
\begin{figure}[htb]
  \centering
  \includegraphics[width=1.0\linewidth]{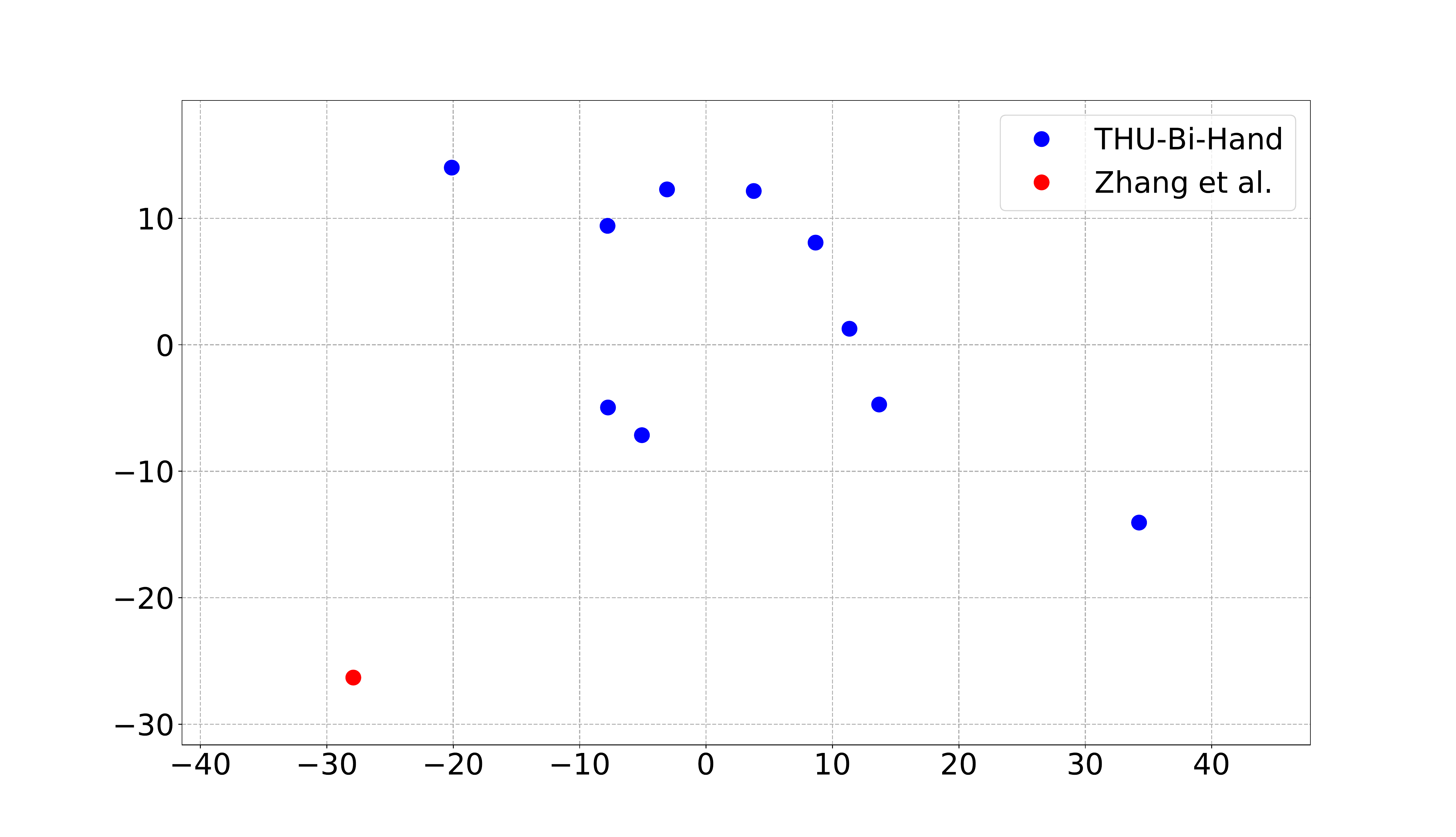}
  \caption{2D PCA visualization of hand shapes. THU-Bi-Hand contains 10 hand shapes while Zhang~et~al.~\cite{zhang2017hand} includes only one.}
  \label{fig:shape}
\end{figure}

There are totally 10 different hand shapes in the THU-Bi-Hand dataset, while~\cite{zhang2017hand} only contains one hand shape. Fig.~\ref{fig:shape} shows the 2D PCA projections of hand shapes in these two datasets.

\section{Experiments and Discussions}
\label{sec:experiment}
In this section, we first evaluate our Bi-Pose-REN framework as well as Chen~et~al.~\cite{chen2016accurate} and TSBNet~\cite{wei2017two} on the ThuHand17 and THU-Bi-Hand datasets as benchmarks for further research on fingertip localization from stereo images. Furthermore, we conduct extra experiments on the THU-Bi-Hand dataset to discuss the effectiveness of different modules in Bi-Pose-REN.

\subsection{Experimental Setup}
\label{sec:setup}
We implemented our Bi-Pose-REN with Caffe~\cite{jia2014caffe} using C++. We used stochastic gradient descent (SGD) with a mini-batch size of 128. For experiments on ThuHand17 dataset, the learning rate started from 0.001 and was divided by 10 on iteration 100k, 160k and 200k, and the model was trained for total 240k iterations. As for THU-Bi-Hand dataset, the learning rate was divided by 10 on iterations 300k, 500k and 600k, and the model was trained for total 700k iterations. Besides, we used a weight decay of 0.0005 and a momentum of 0.9. For Bi-Pose-REN, we trained the model for two iterations, and used the final model of the second iteration to test for one iteration.

\subsection{Evaluation Metrics}
\label{sec:metrics}
Following~\cite{wei2017two}, the performance is evaluated by two metrics: 1) \emph{average 3D distance error} is computed as the Euclidean distance between the ground truths and the predictions of 3D coordinates in the coordinate system of Leap Motion (in millimeters). The mean of average 3D distance errors of all the six joints is also presented; 2) \emph{percentage of success frames} is defined as the percentage of frames in which all 3D Euclidean errors of joints are below a threshold.

\subsection{Benchmarks of Fingertip Localization}
\label{sec:comparison}
\begin{figure*}[htb]
  \begin{minipage}[b]{0.50\textwidth}
    \centering
    \centerline{\includegraphics[width=0.99\textwidth]{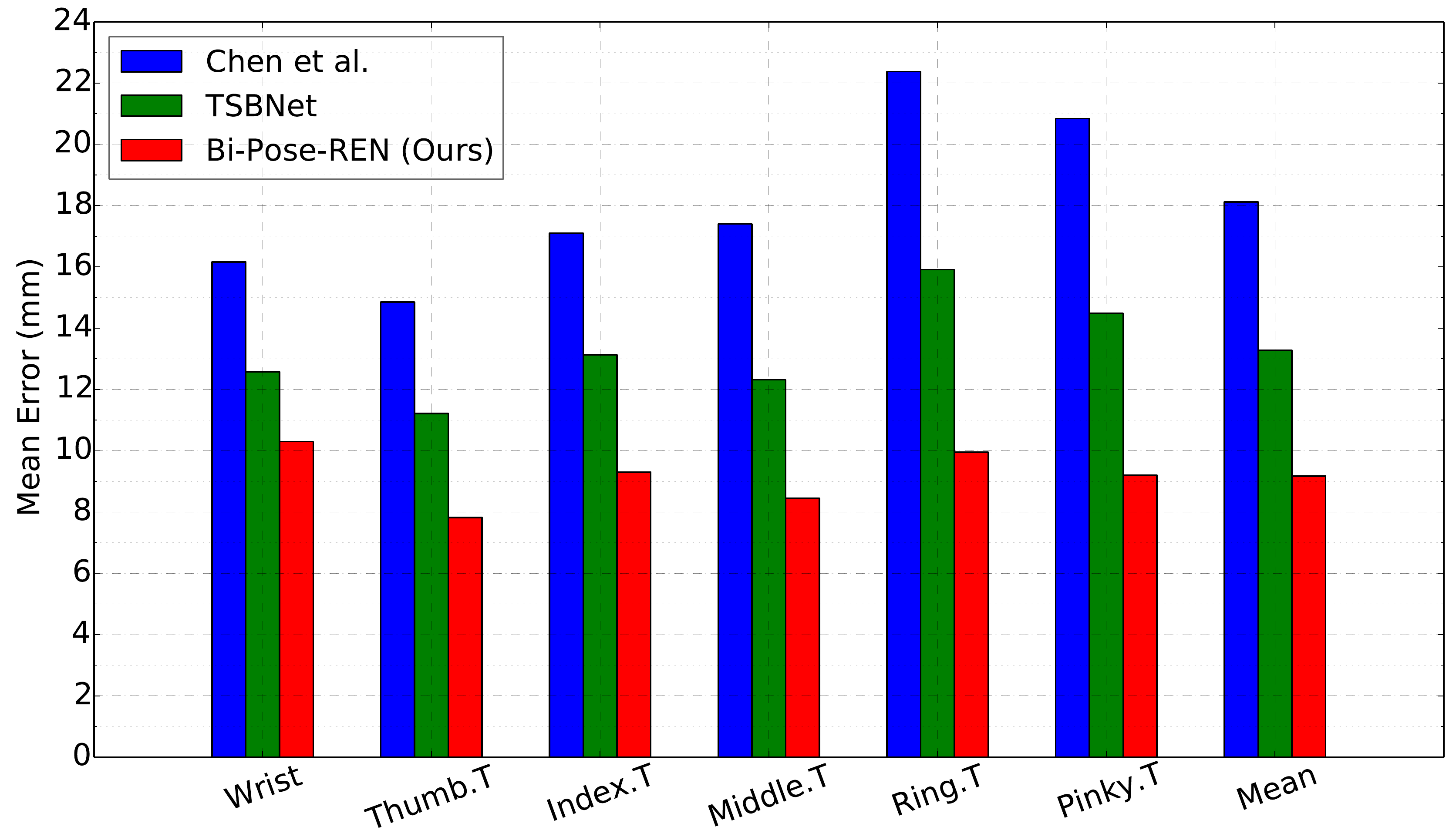}}
  \end{minipage}
  \begin{minipage}[b]{0.49\textwidth}
    \centering
    \centerline{\includegraphics[width=0.99\textwidth]{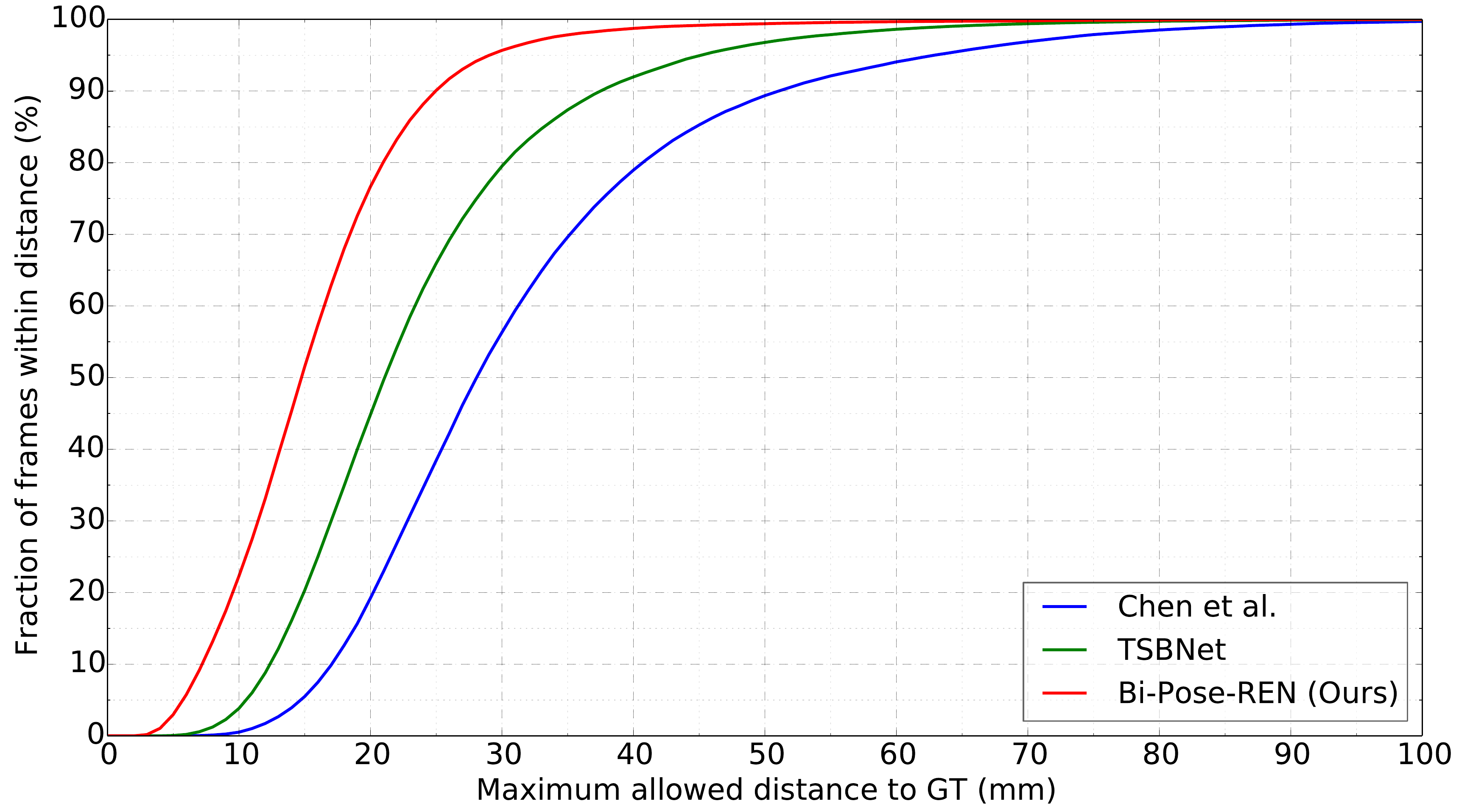}}
  \end{minipage}
  \caption{Effect of Bi-Pose-REN on the THU-Bi-Hand dataset compared with Chen~et~al.~\cite{chen2016accurate} and TSBNet~\cite{wei2017two}. Left: average 3D distance error. Right: percentage of success frames.}
  \label{fig:comparison_thubihand}
\end{figure*}

To demonstrate the effectiveness of our Bi-Pose-REN, we compare it against previous work, noted as Chen~et~al.~\cite{chen2016accurate} and TSBNet~\cite{wei2017two}. The average 3D distance error and the percentage of success frames on the THU-Bi-Hand dataset are shown in Fig.~\ref{fig:comparison_thubihand}. The quantitative results of the mean error of all the joints (the rightmost three bars labelled as ``$Mean$'') on both datasets are presented in Table~\ref{table:comparison}.

\begin{table}
  \renewcommand{\arraystretch}{1.3}
  \caption{Comparison with state-of-the-arts on the ThuHand17 and THU-Bi-Hand datasets. Our Bi-Pose-REN outperforms others.}
  \label{table:comparison}
  \centering
  \begin{tabular}{ccc}
    \hline
    Method & ThuHand17 & THU-Bi-Hand\\
    \hline
    Chen~et~al.~\cite{chen2016accurate} & 16.84mm & 18.12mm \\
    TSBNet~\cite{wei2017two} & 10.91mm & 13.27mm \\
    \textbf{Bi-Pose-REN (Ours)} & \textbf{8.08mm} & \textbf{9.17mm} \\
    \hline
  \end{tabular}
\end{table}

Compared with Chen~et~al.~\cite{chen2016accurate} and TSBNet~\cite{wei2017two}, the mean errors reduced from 16.84mm and 10.91mm to 8.08mm (with about 26\% improvement in contrast with TSBNet) on the ThuHand17 dataset. As for THU-Bi-Hand, the mean error of Bi-Pose-REN outperforms Chen~et~al.~\cite{chen2016accurate} and TSBNet~\cite{wei2017two} by 49.39\% and 30.90\% respectively. As shown in Fig.~\ref{fig:comparison_thubihand}, the average errors of each joint estimated with Bi-Pose-REN are much smaller than Chen~et~al.~\cite{chen2016accurate} and TSBNet~\cite{wei2017two} on THU-Bi-Hand dataset. Except the wrist, all the other joints have their average errors smaller than 10mm for Bi-Pose-REN.

The percentage of success frames of Bi-Pose-REN is better than Chen~et~al.~\cite{chen2016accurate} and TSBNet~\cite{wei2017two} consistently, no matter with a small or large threshold. Specifically on the THU-Bi-Hand dataset, under a threshold of 20mm, Bi-Pose-REN successfully predicts more than 75\% of frames, while TSBNet~\cite{wei2017two} about 45\% and Chen~et~al.~\cite{chen2016accurate} less than 20\%. In summary, Bi-Pose-REN not only predicts more accurate positions of fingertips and wrist than previous methods, but also outperforms others in providing high-quality frame predictions under different levels of requirements of accuracy.

Bi-Pose-REN runs at 55fps on an NVIDIA GeForce 1080TI GPU during inference (6.6ms for initialization plus 11.6ms for refinement), which is promising for real-time applications.

\subsection{Ablation Studies}
\label{sec:ablation}
To study the ablation of our Bi-Pose-REN, we first introduce the modules and then evaluate different methods generated by concatenating different modules.

\subsubsection{Module Introduction}
\label{sec:intro_module}
\ \par
\begin{figure*}[htb]
  \centering
  \includegraphics[width=1.0\textwidth]{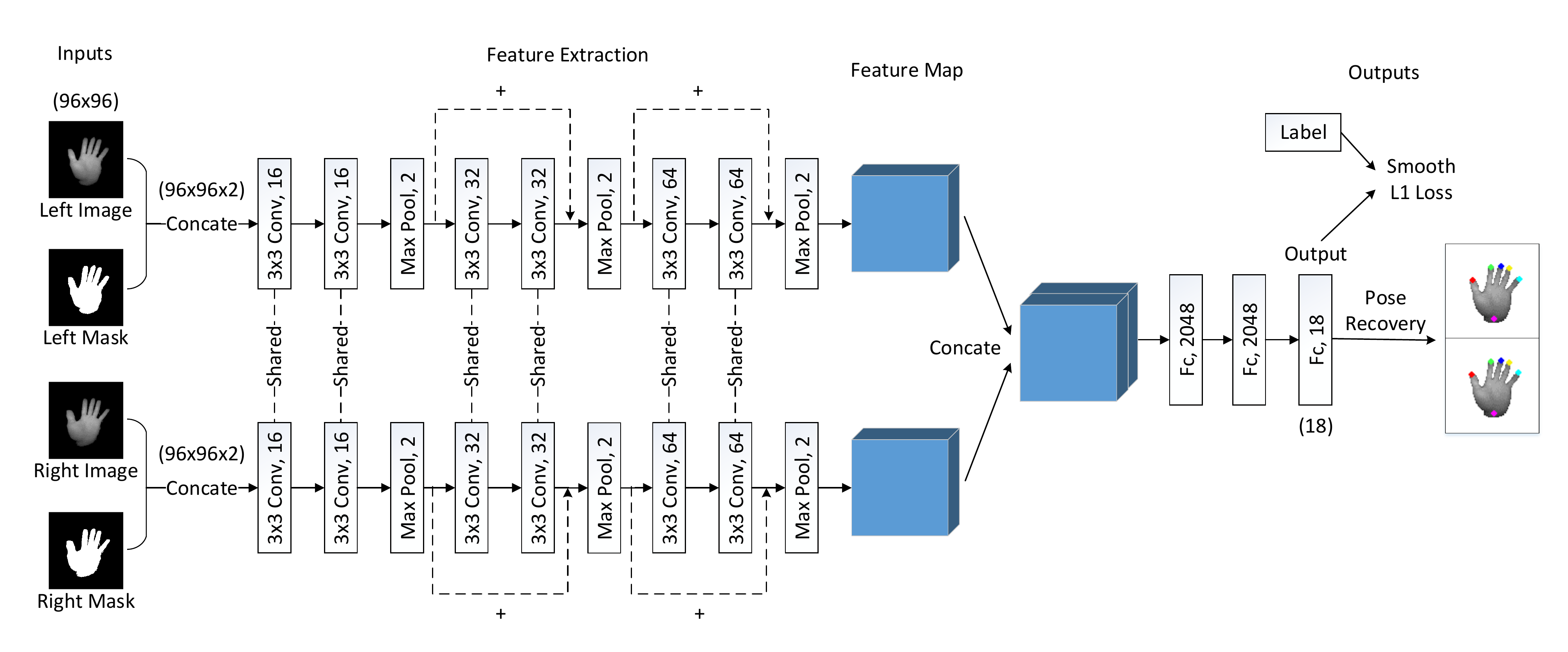}\\
  \caption{The basic CNN architecture in two streams. ``$k \times k \ Conv, c$'' represents a convolutional layer with a $k \times k$ size kernel and $c$ channels ($stride = 1, pad = 1$). ``$Max \ Pool, 2$'' denotes a max pooling layer with kernel size $2 \times 2 $ ($stride = 2, no \ padding$). ``$Fc, n$'' represents a fully connected layer with $n$ neurons. The numbers inside parentheses denote the sizes of images or vectors.}
  \label{fig:base_feat}
\end{figure*}

\textbf{Feature Extraction Module.} In Bi-Pose-REN, we use the structure of DenseNet~\cite{huang2017densely} to extract feature maps from stereo images. For comparison, a basic CNN architecture (the baseline in~\cite{wang2018region} with residual connections, but in the two-stream style for stereo images) is used for feature extraction in other methods, see Fig.~\ref{fig:base_feat}. Average pooling is used in DenseNet as~\cite{huang2017densely}, while max pooling is used in the basic CNN, following~\cite{wang2018region}.

\textbf{Regression Module.} Region ensemble methods are proved to be effective in hand pose estimation from depth images~\cite{guo2017region,wang2018region,chen2017pose}. To demonstrate the performances of different models on the THU-Bi-Hand dataset in the best way, we employ different region ensemble methods for fingertip localization from stereo images. REN is used in some methods while Pose-REN in others. While using REN, as in~\cite{wang2018region}, nine regions are extracted in each stream respectively. The regions of the left and right streams are concatenated and fed into two FC layers. Afterwards, the nine outputs are fused by concatenation and regress the hand pose.

By concatenating different modules, we present several methods, see Table~\ref{table:methods}.

\begin{table}[htb]
  \renewcommand{\arraystretch}{1.3}
  \caption{The different methods for fingertip localization from stereo images. Feature extraction modules of all methods are in two same-structure and shared-parameter streams.}
  \label{table:methods}
  \centering
  \begin{tabular}{ccc}
    \hline
    Method & Feature Extraction & Regression Architecture \\
    \hline
    Basic-CNN &  Basic CNN & FC layers \\
    DenseNet-CNN &  DenseNet & FC layers \\
    DenseNet-REN & DenseNet & REN \\
    Bi-Pose-REN & DenseNet & Pose-REN \\
    \hline
  \end{tabular}
\end{table}

\subsubsection{Module Effects}
\label{sec:module_effects}
\ \par
\begin{figure*}[htb]
  \begin{minipage}[b]{0.50\textwidth}
    \centering
    \centerline{\includegraphics[width=0.99\textwidth]{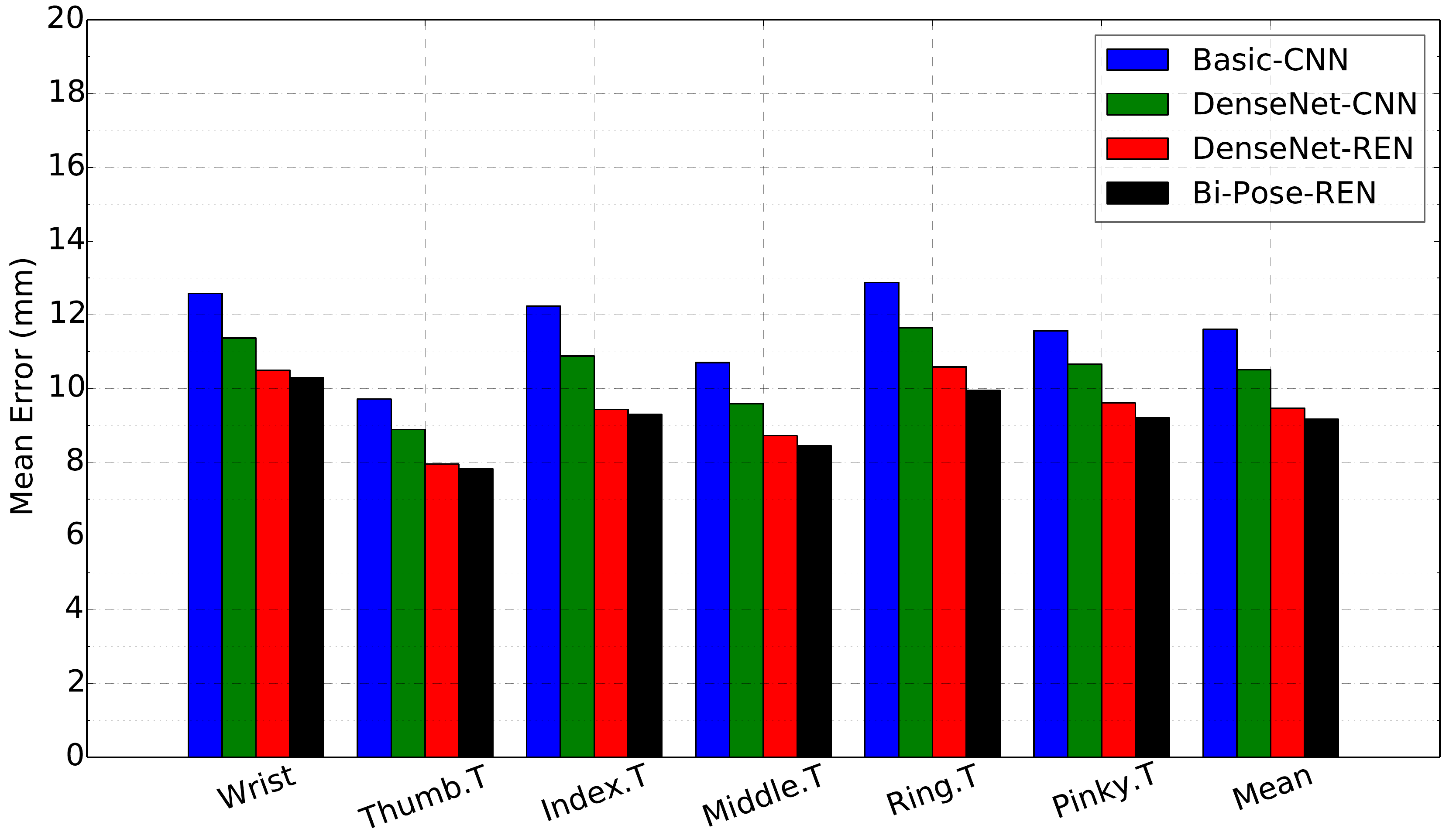}}
  \end{minipage}
  \begin{minipage}[b]{0.49\textwidth}
    \centering
    \centerline{\includegraphics[width=0.99\textwidth]{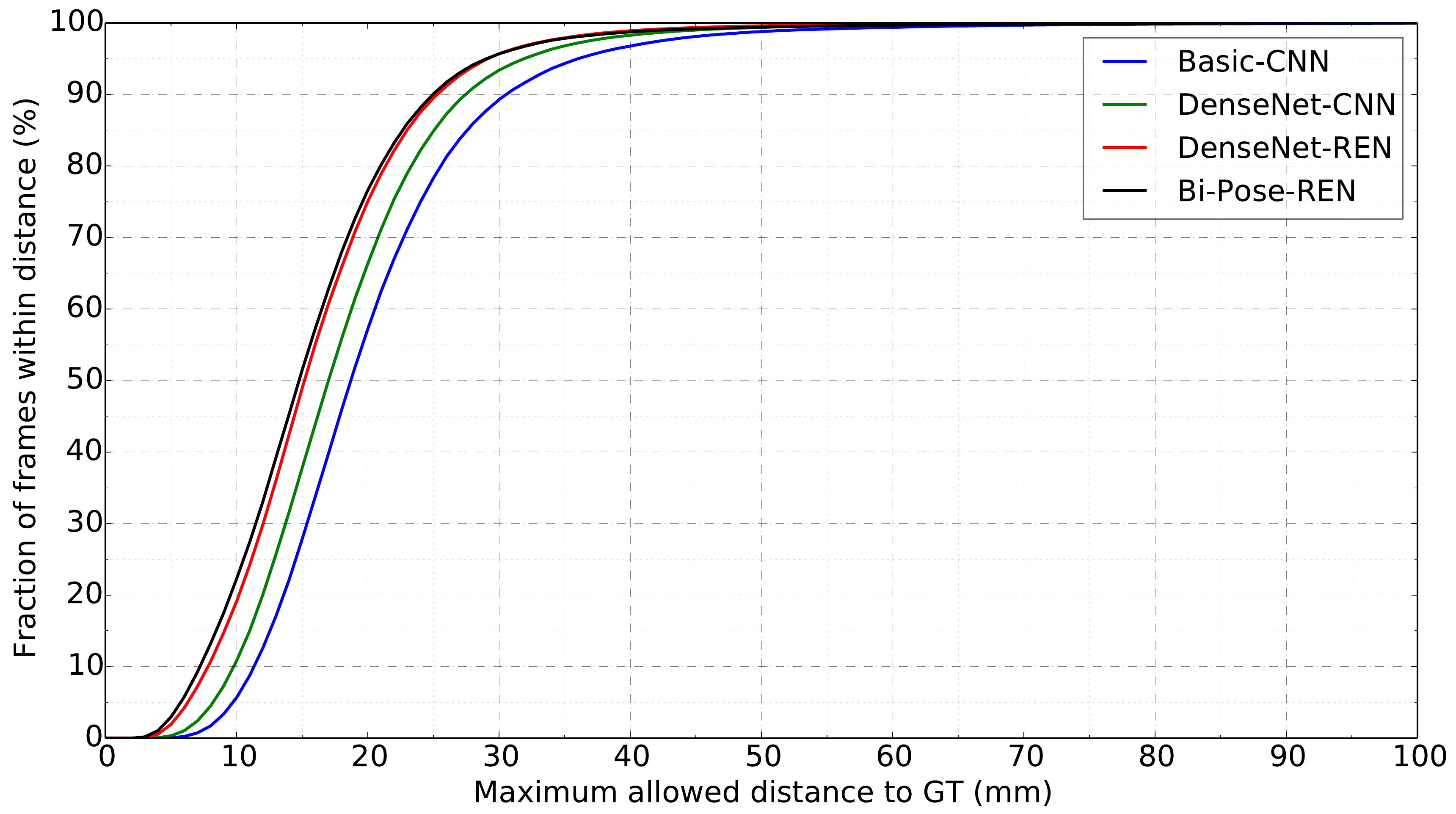}}
  \end{minipage}
  \caption{Module effects of Bi-Pose-REN on THU-Bi-Hand dataset. Left: average 3D distance error. Right: percentage of success frames.}
  \label{fig:ablation}
\end{figure*}

\textbf{Feature Extraction.} We test two kinds of feature extraction modules: a shallow CNN in two streams with the same structure and shared parameters and residual connections, and a two-stream DenseNet structured CNN. Compared with Basic-CNN, DenseNet-CNN produces a smaller mean error for all the six joints (See Fig.~\ref{fig:ablation}). The mean error of all the joints is 10.51mm by using DenseNet-CNN (9.47\% better than Basic-CNN, see Table~\ref{table:ablation}). The percentage of success frames is also higher by the DenseNet structure for feature extraction.

\textbf{Regression Module.} Without region ensemble, DenseNet-CNN performs poorer than DenseNet-REN in per joint error (10.51mm versus 9.47mm, see Table~\ref{table:ablation} and Fig.~\ref{fig:ablation}, DenseNet-REN is 9.90\% better). In the meanwhile, Bi-Pose-REN outperforms DenseNet-REN by 3.17\%. That is, the pose guided structured region ensemble method beats the region ensemble network without guidance, while it performs the worst without region ensemble. Similar patterns can be observed on the ThuHand17 dataset.

\begin{table}
  \renewcommand{\arraystretch}{1.3}
  \centering
  \caption{Ablation studies of Bi-Pose-REN on THU-Bi-Hand and ThuHand17 datasets.}
  \label{table:ablation}
  \begin{tabular}{ccc}
    \hline
    Method & THU-Bi-Hand & ThuHand17 \\
    \hline
    Basic-CNN & 11.61mm & 10.76mm \\
    DenseNet-CNN & 10.51mm & 10.20mm \\
    DenseNet-REN & 9.47mm & 8.98mm \\
    \textbf{Bi-Pose-REN} & \textbf{9.17mm} & \textbf{8.08mm} \\
    \hline
  \end{tabular}
\end{table}

\begin{figure*}[htb]
  \begin{minipage}[b]{0.50\textwidth}
    \centering
    \centerline{\includegraphics[width=0.99\textwidth]{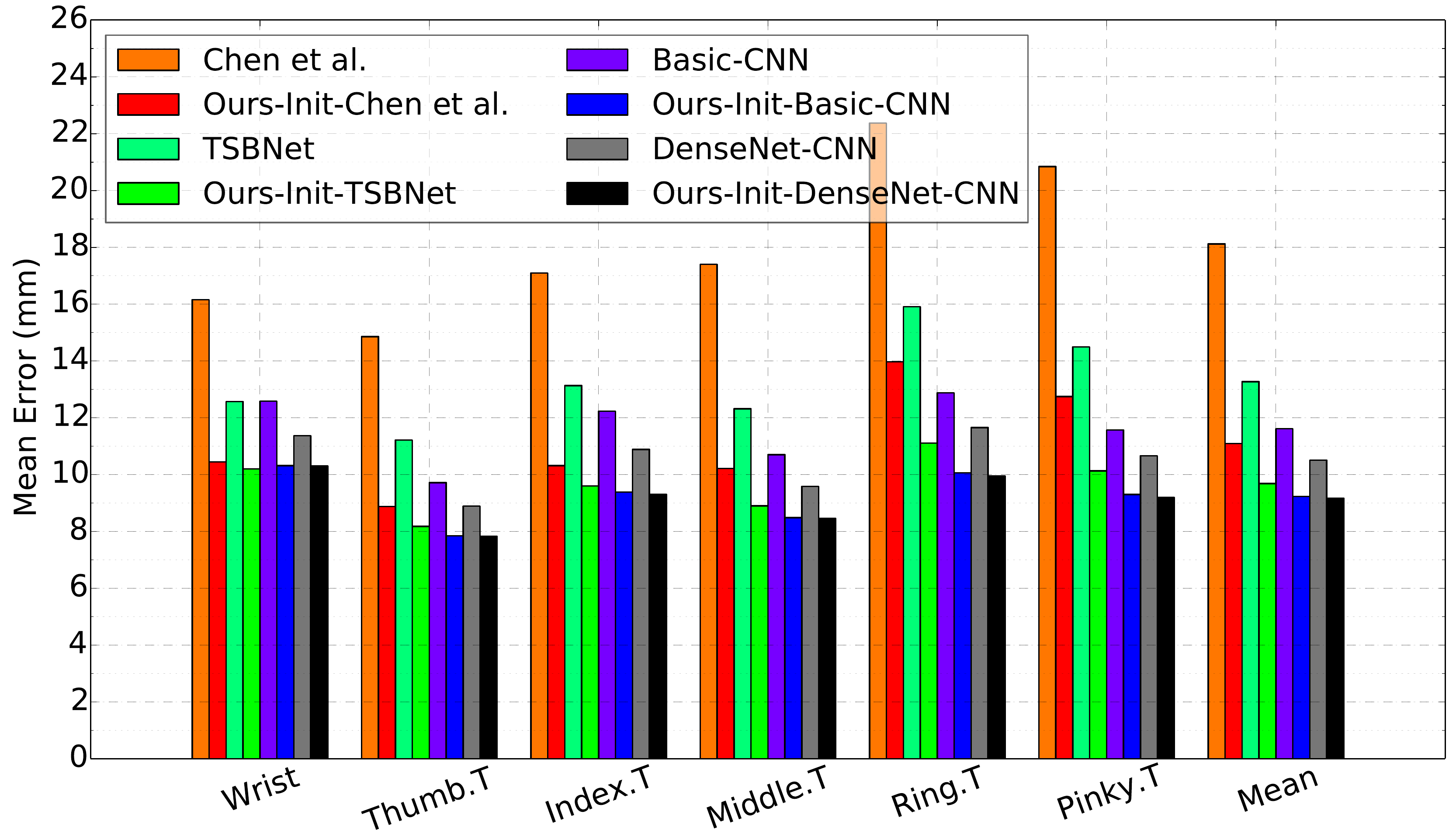}}
  \end{minipage}
  \begin{minipage}[b]{0.49\textwidth}
    \centering
    \centerline{\includegraphics[width=0.99\textwidth]{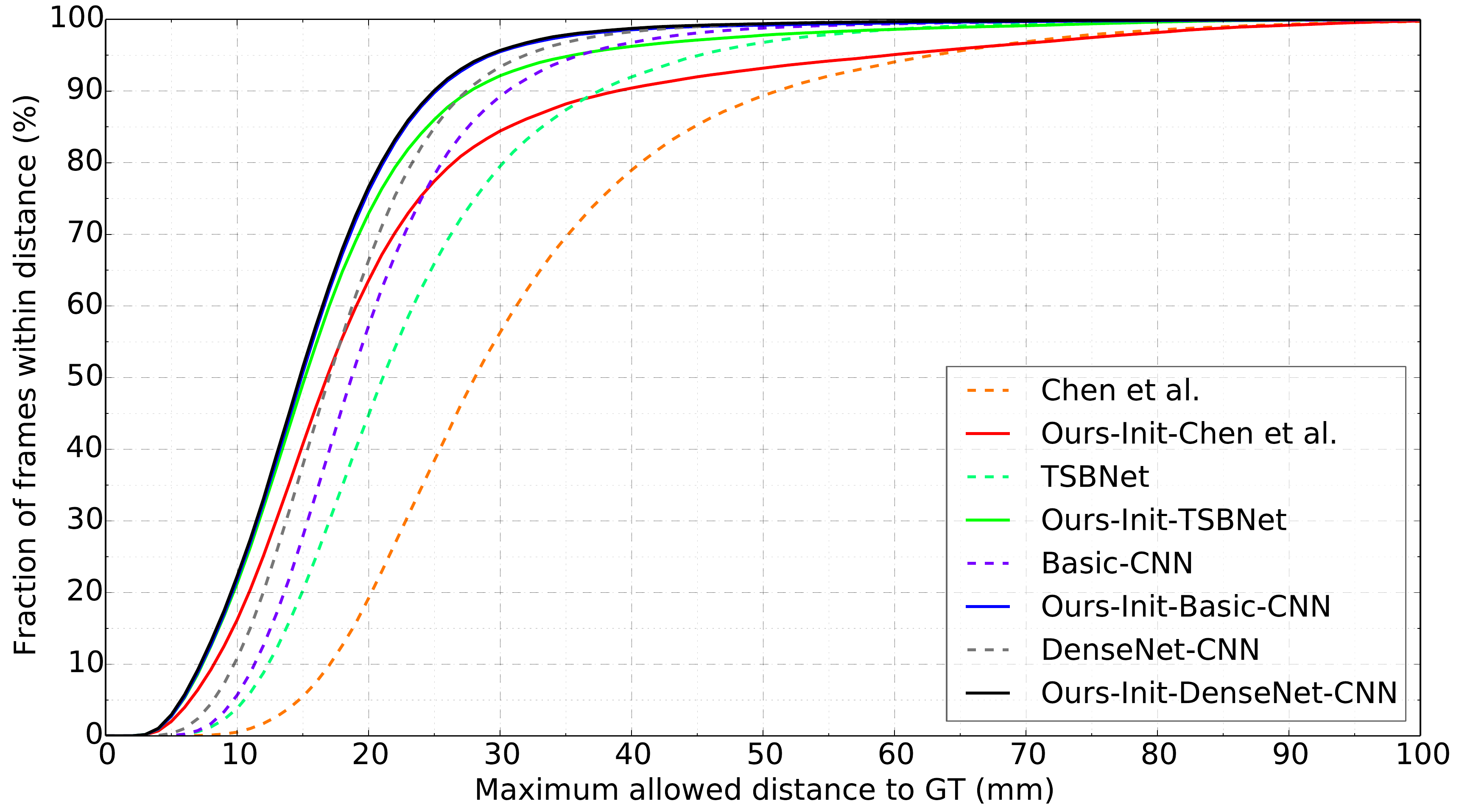}}
  \end{minipage}
  \caption{Effects of different initialization methods of Bi-Pose-REN on the THU-Bi-Hand dataset. Left: average 3D distance error. Right: percentage of success frames.}
  \label{fig:initialize}
\end{figure*}

\textbf{Initialization Network.} We explore the robustness of Bi-Pose-REN over different pose initializations during inference. Different methods including Basic-CNN, DenseNet-CNN, Chen~et~al.~\cite{chen2016accurate} and TSBNet~\cite{wei2017two} were used to provide the initial pose in inference phase. The results of different initializations and refined results are shown in Fig.~\ref{fig:initialize} and Table~\ref{table:initialize}. It can be seen that our method boosts the performances of initializations. Even with some rather poor initializations (Chen~et~al. and TSBNet), the refined results are quite competitive. With better initializations (Basic-CNN and DenseNet-CNN), the final results are similar. The results demonstrate the robustness over initializations of our Bi-Pose-REN. Note that the model used above was trained by using the samples with DenseNet-CNN initialization.

\begin{table}
  \renewcommand{\arraystretch}{1.3}
  \centering
  \caption{Effects of different initialization methods of Bi-Pose-REN on THU-Bi-Hand and ThuHand17 datasets. The average 3D distance errors in millimeters of refined and initial poses are shown outside and inside parentheses respectively.}
  \label{table:initialize}
  \begin{tabular}{ccc}
    \hline
    Init Net & THU-Bi-Hand & ThuHand17 \\
    \hline
    Chen~et~al.~\cite{chen2016accurate} & 11.10 (18.12) & 9.80 (16.84) \\
    TSBNet~\cite{wei2017two} & 9.69 (13.27) & 8.41 (10.91) \\
    Basic-CNN & 9.23 (11.61) & 8.25 (10.76) \\
    \textbf{DenseNet-CNN} & \textbf{9.17 (10.51)} & \textbf{8.08 (10.20)} \\
    \hline
  \end{tabular}
\end{table}

\begin{figure}[htb]
  \centering
  \includegraphics[width=1.0\columnwidth]{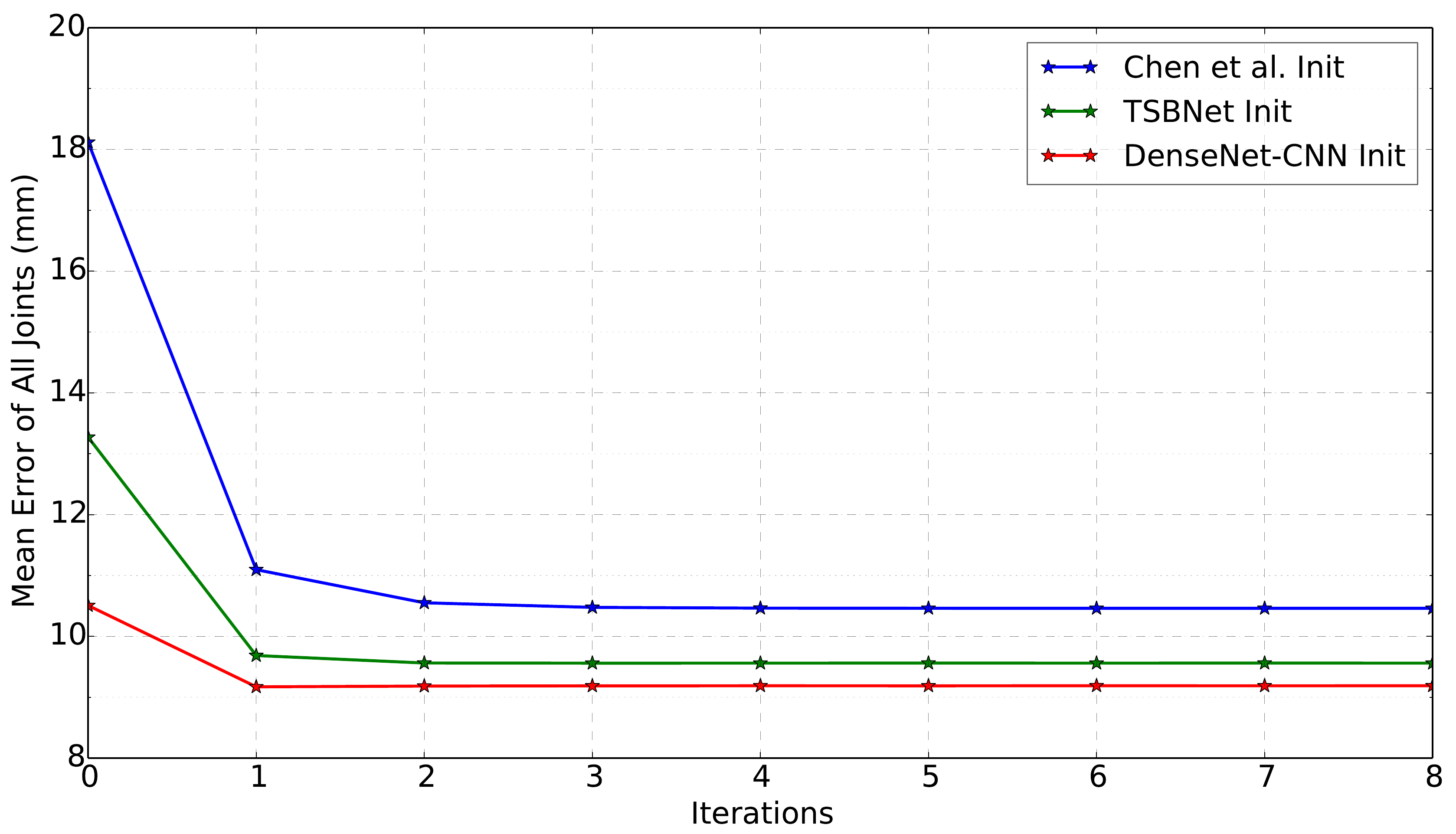}
  \caption{Effect of the number of iterations in Bi-Pose-REN.}
  \label{fig:iter}
\end{figure}

\textbf{Iterations in Pose-REN.} Bi-Pose-REN are cascaded frameworks with iterations. We explore the effect of the number of iterations by iteratively testing the Bi-Pose-REN model. The model was trained by using the samples with DenseNet-CNN initialization. The results of refined mean error of all the joints obtained from using different Init Nets during inference are presented in Fig.~\ref{fig:iter}. It can be seen that Bi-Pose-REN converges very fast, as the results after only one or two iterations are adorable.

\subsection{Qualitative Results}
\label{sec:qualititive}
\begin{figure*}[!htb]
  \centering
  \includegraphics[width=0.98\textwidth]{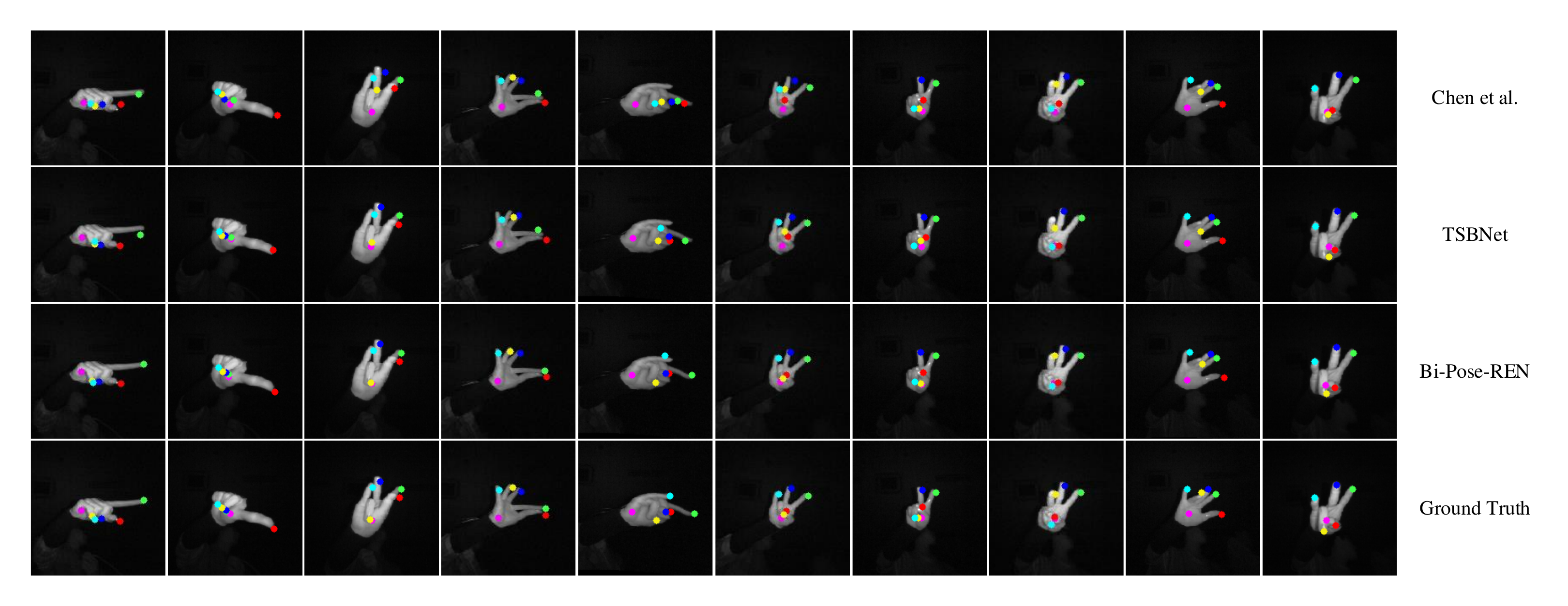}
  \caption{Qualitative results on the THU-Bi-Hand dataset. Predictions of Chen~et~al.~\cite{chen2016accurate}, TSBNet~\cite{wei2017two} and our Bi-Pose-REN as well as the ground truths are shown in different rows. Bi-Pose-REN predicts promising results even in some difficult cases.}
  \label{fig:samples}
\end{figure*}

Fig.~\ref{fig:samples} shows some examples of the results of Bi-Pose-REN as well as Chen~et~al.~\cite{chen2016accurate} and TSBNet~\cite{wei2017two} evaluated on the THU-Bi-Hand dataset. The predictions of Bi-Pose-REN are very closed to the ground truths. The estimation is promising even for some difficult cases, such as several fingers occluded and side viewpoints, while Chen~et~al. and TSBNet perform poorer in these cases.

\section{Conclusion}
\label{sec:conclusion}
In this paper we proposed a large binocular hand pose dataset called THU-Bi-Hand, which contains about 447k frames of stereo images with accurate 3D location annotations of the wrist and five fingertips captured from 10 different subjects. THU-Bi-Hand covers large diversity of global viewpoints, hand shapes, hand poses and hand articulations, which has potential benefits to human-computer interaction with hand poses.
We also proposed a novel scheme termed Bi-Pose-REN for fingertip localization on THU-Bi-Hand. Feature regions are cropped from the feature maps around each joint position given an initial hand pose and fused by FC layers to regress a refined pose iteratively.
We evaluated several methods including Bi-Pose-REN on the THU-Bi-Hand dataset to provide benchmarks for promoting further research on fingertip localization from stereo images. Hand pose estimation from stereo images directly is then practicable with high accuracy.
Future work will focus on fingertip localization from noisy stereo images or monocular RGB images, as well as hand pose estimation with hand-object interaction.


%





\ifCLASSOPTIONcaptionsoff
  \newpage
\fi



\bibliographystyle{IEEEtran}
\bibliography{IEEEabrv,refs}

\begin{thebibliography}{10}
\providecommand{\url}[1]{#1}
\csname url@samestyle\endcsname
\providecommand{\newblock}{\relax}
\providecommand{\bibinfo}[2]{#2}
\providecommand{\BIBentrySTDinterwordspacing}{\spaceskip=0pt\relax}
\providecommand{\BIBentryALTinterwordstretchfactor}{4}
\providecommand{\BIBentryALTinterwordspacing}{\spaceskip=\fontdimen2\font plus
\BIBentryALTinterwordstretchfactor\fontdimen3\font minus
  \fontdimen4\font\relax}
\providecommand{\BIBforeignlanguage}[2]{{%
\expandafter\ifx\csname l@#1\endcsname\relax
\typeout{** WARNING: IEEEtran.bst: No hyphenation pattern has been}%
\typeout{** loaded for the language `#1'. Using the pattern for}%
\typeout{** the default language instead.}%
\else
\language=\csname l@#1\endcsname
\fi
#2}}
\providecommand{\BIBdecl}{\relax}
\BIBdecl

\bibitem{erol2007vision}
A.~Erol, G.~Bebis, M.~Nicolescu, R.~D. Boyle, and X.~Twombly, ``Vision-based
  hand pose estimation: A review,'' \emph{Computer Vision and Image
  Understanding}, vol. 108, no. 1-2, pp. 52--73, 2007.

\bibitem{lee2014real}
J.~H. Lee, T.~Delbruck, M.~Pfeiffer, P.~K. Park, C.-W. Shin, H.~Ryu, and B.~C.
  Kang, ``Real-time gesture interface based on event-driven processing from
  stereo silicon retinas,'' \emph{IEEE Transactions on Neural Networks and
  Learning Systems}, vol.~25, no.~12, pp. 2250--2263, 2014.

\bibitem{nolker2002visual}
C.~Nolker and H.~Ritter, ``Visual recognition of continuous hand postures,''
  \emph{IEEE Transactions on neural networks}, vol.~13, no.~4, pp. 983--994,
  2002.

\bibitem{zhao2015learning}
L.~Zhao, X.~Gao, D.~Tao, and X.~Li, ``Learning a tracking and estimation
  integrated graphical model for human pose tracking,'' \emph{IEEE Transactions
  on Neural Networks and Learning Systems}, vol.~26, no.~12, pp. 3176--3186,
  2015.

\bibitem{tompson2014real}
J.~Tompson, M.~Stein, Y.~Lecun, and K.~Perlin, ``Real-time continuous pose
  recovery of human hands using convolutional networks,'' \emph{ACM
  Transactions on Graphics (ToG)}, vol.~33, no.~5, p. 169, 2014.

\bibitem{supancic2015depth}
J.~S. Supancic, G.~Rogez, Y.~Yang, J.~Shotton, and D.~Ramanan, ``Depth-based
  hand pose estimation: data, methods, and challenges,'' in \emph{Proceedings
  of the IEEE international conference on computer vision}, 2015, pp.
  1868--1876.

\bibitem{tang2015opening}
D.~Tang, J.~Taylor, P.~Kohli, C.~Keskin, T.-K. Kim, and J.~Shotton, ``Opening
  the black box: Hierarchical sampling optimization for estimating human hand
  pose,'' in \emph{Proceedings of the IEEE International Conference on Computer
  Vision}, 2015, pp. 3325--3333.

\bibitem{oberweger2015training}
M.~Oberweger, P.~Wohlhart, and V.~Lepetit, ``Training a feedback loop for hand
  pose estimation,'' in \emph{Proceedings of the IEEE International Conference
  on Computer Vision}, 2015, pp. 3316--3324.

\bibitem{wan2016hand}
C.~Wan, A.~Yao, and L.~Van~Gool, ``Hand pose estimation from local surface
  normals,'' in \emph{European Conference on Computer Vision}.\hskip 1em plus
  0.5em minus 0.4em\relax Springer, 2016, pp. 554--569.

\bibitem{wan2017crossing}
C.~Wan, T.~Probst, L.~Van~Gool, and A.~Yao, ``Crossing nets: Combining {GANs}
  and {VAEs} with a shared latent space for hand pose estimation,'' in
  \emph{2017 IEEE Conference on Computer Vision and Pattern Recognition
  (CVPR)}.\hskip 1em plus 0.5em minus 0.4em\relax IEEE, 2017.

\bibitem{yuan2018depth}
S.~Yuan, G.~Garcia-Hernando, B.~Stenger, G.~Moon, J.~Y. Chang, K.~M. Lee,
  P.~Molchanov, J.~Kautz, S.~Honari, L.~Ge \emph{et~al.}, ``Depth-based {3D}
  hand pose estimation: From current achievements to future goals,'' in
  \emph{IEEE CVPR}, 2018.

\bibitem{moon2018v2v}
G.~Moon, J.~Y. Chang, and K.~M. Lee, ``{V2V-PoseNet}: Voxel-to-voxel prediction
  network for accurate {3D} hand and human pose estimation from a single depth
  map,'' in \emph{CVPR}, vol.~2, no.~3, 2018.

\bibitem{ge20173d}
L.~Ge, H.~Liang, J.~Yuan, and D.~Thalmann, ``{3D} convolutional neural networks
  for efficient and robust hand pose estimation from single depth images,'' in
  \emph{Proceedings of the IEEE Conference on Computer Vision and Pattern
  Recognition}, vol.~1, 2017, p.~5.

\bibitem{zhang2018interactive}
C.~Zhang, G.~Wang, H.~Guo, X.~Chen, F.~Qiao, and H.~Yang, ``Interactive hand
  pose estimation: Boosting accuracy in localizing extended finger joints,''
  \emph{Electronic Imaging}, vol. 2018, no.~2, pp. 251--1--251--6, 2018.

\bibitem{zhang2012microsoft}
Z.~Zhang, ``Microsoft {Kinect} sensor and its effect,'' \emph{IEEE multimedia},
  vol.~19, no.~2, pp. 4--10, 2012.

\bibitem{wang2013depth}
G.~Wang, X.~Yin, X.~Pei, and C.~Shi, ``Depth estimation for speckle projection
  system using progressive reliable points growing matching,'' \emph{Applied
  optics}, vol.~52, no.~3, pp. 516--524, 2013.

\bibitem{shi2015high}
C.~Shi, G.~Wang, X.~Yin, X.~Pei, B.~He, and X.~Lin, ``High-accuracy stereo
  matching based on adaptive ground control points,'' \emph{IEEE Transactions
  on Image Processing}, vol.~24, no.~4, pp. 1412--1423, 2015.

\bibitem{keselman2017intel}
L.~Keselman, J.~Iselin~Woodfill, A.~Grunnet-Jepsen, and A.~Bhowmik, ``Intel
  {RealSense} stereoscopic depth cameras,'' in \emph{Proceedings of the IEEE
  Conference on Computer Vision and Pattern Recognition Workshops}, 2017, pp.
  1--10.

\bibitem{rogister2012asynchronous}
P.~Rogister, R.~Benosman, S.-H. Ieng, P.~Lichtsteiner, and T.~Delbruck,
  ``Asynchronous event-based binocular stereo matching,'' \emph{IEEE
  Transactions on Neural Networks and Learning Systems}, vol.~23, no.~2, pp.
  347--353, 2012.

\bibitem{chen2016accurate}
X.~Chen, G.~Wang, and H.~Guo, ``Accurate fingertip detection from binocular
  mask images,'' in \emph{Visual Communications and Image Processing (VCIP),
  2016}.\hskip 1em plus 0.5em minus 0.4em\relax IEEE, 2016, pp. 1--4.

\bibitem{wei2017two}
Y.~Wei, G.~Wang, C.~Zhang, H.~Guo, X.~Chen, and H.~Yang, ``Two-stream binocular
  network: Accurate near field finger detection based on binocular images,'' in
  \emph{Visual Communications and Image Processing (VCIP), 2017 IEEE}.\hskip
  1em plus 0.5em minus 0.4em\relax IEEE, 2017, pp. 1--4.

\bibitem{zhang2017hand}
J.~Zhang, J.~Jiao, M.~Chen, L.~Qu, X.~Xu, and Q.~Yang, ``A hand pose tracking
  benchmark from stereo matching,'' in \emph{24th IEEE International Conference
  on Image Processing, ICIP 2017}, 2017.

\bibitem{huang2017densely}
G.~Huang, Z.~Liu, K.~Q. Weinberger, and L.~van~der Maaten, ``Densely connected
  convolutional networks,'' in \emph{Proceedings of the IEEE conference on
  computer vision and pattern recognition}, vol.~1, no.~2, 2017, p.~3.

\bibitem{tang2014latent}
D.~Tang, H.~Jin~Chang, A.~Tejani, and T.-K. Kim, ``Latent regression forest:
  Structured estimation of {3D} articulated hand posture,'' in
  \emph{Proceedings of the IEEE Conference on Computer Vision and Pattern
  Recognition}, 2014, pp. 3786--3793.

\bibitem{sun2015cascaded}
X.~Sun, Y.~Wei, S.~Liang, X.~Tang, and J.~Sun, ``Cascaded hand pose
  regression,'' in \emph{Proceedings of the IEEE Conference on Computer Vision
  and Pattern Recognition}, 2015, pp. 824--832.

\bibitem{yuan2017bighand2}
S.~Yuan, Q.~Ye, B.~Stenger, S.~Jain, and T.-K. Kim, ``{BigHand2.2M} benchmark:
  Hand pose dataset and state of the art analysis,'' in \emph{Computer Vision
  and Pattern Recognition (CVPR), 2017 IEEE Conference on}.\hskip 1em plus
  0.5em minus 0.4em\relax IEEE, 2017, pp. 2605--2613.

\bibitem{wetzler2015rule}
A.~Wetzler, R.~Slossberg, and R.~Kimmel, ``Rule of thumb: Deep derotation for
  improved fingertip detection,'' \emph{arXiv preprint arXiv:1507.05726}, 2015.

\bibitem{oberweger2015hands}
M.~Oberweger, P.~Wohlhart, and V.~Lepetit, ``Hands deep in deep learning for
  hand pose estimation,'' in \emph{Proceedings of Computer Vision Winter
  Workshop}, 2015, pp. 21--30.

\bibitem{basaru2017hand}
R.~R. Basaru, C.~Child, E.~Alonso, and G.~Slabaugh, ``Hand pose estimation
  using deep stereovision and markov-chain monte carlo,'' in \emph{2017 IEEE
  International Conference on Computer Vision Workshop (ICCVW)}.\hskip 1em plus
  0.5em minus 0.4em\relax IEEE, 2017, pp. 595--603.

\bibitem{romero2008dynamic}
J.~Romero, D.~Kragic, V.~Kyrki, and A.~Argyros, ``Dynamic time warping for
  binocular hand tracking and reconstruction,'' in \emph{Robotics and
  Automation, 2008. ICRA 2008. IEEE International Conference on}.\hskip 1em
  plus 0.5em minus 0.4em\relax IEEE, 2008, pp. 2289--2294.

\bibitem{panteleris2017back}
P.~Panteleris and A.~Argyros, ``Back to rgb: 3d tracking of hands and
  hand-object interactions based on short-baseline stereo,'' in \emph{2017 IEEE
  International Conference on Computer Vision Workshop (ICCVW)}.\hskip 1em plus
  0.5em minus 0.4em\relax IEEE, 2017, pp. 575--584.

\bibitem{qian2014realtime}
C.~Qian, X.~Sun, Y.~Wei, X.~Tang, and J.~Sun, ``Realtime and robust hand
  tracking from depth,'' in \emph{Proceedings of the IEEE Conference on
  Computer Vision and Pattern Recognition}, 2014, pp. 1106--1113.

\bibitem{oikonomidis2011efficient}
I.~Oikonomidis, N.~Kyriazis, and A.~A. Argyros, ``Efficient model-based {3D}
  tracking of hand articulations using {Kinect},'' in \emph{BMVC}, vol.~1,
  no.~2, 2011, p.~3.

\bibitem{krizhevsky2012imagenet}
A.~Krizhevsky, I.~Sutskever, and G.~E. Hinton, ``{ImageNet} classification with
  deep convolutional neural networks,'' in \emph{Advances in neural information
  processing systems}, 2012, pp. 1097--1105.

\bibitem{simonyan2014very}
K.~Simonyan and A.~Zisserman, ``Very deep convolutional networks for
  large-scale image recognition,'' \emph{arXiv preprint arXiv:1409.1556}, 2014.

\bibitem{szegedy2015going}
C.~Szegedy, W.~Liu, Y.~Jia, P.~Sermanet, S.~Reed, D.~Anguelov, D.~Erhan,
  V.~Vanhoucke, and A.~Rabinovich, ``Going deeper with convolutions,'' in
  \emph{2015 IEEE Conference on Computer Vision and Pattern Recognition
  (CVPR)}.

\bibitem{he2016deep}
K.~He, X.~Zhang, S.~Ren, and J.~Sun, ``Deep residual learning for image
  recognition,'' in \emph{Proceedings of the IEEE conference on computer vision
  and pattern recognition}, 2016, pp. 770--778.

\bibitem{guo2017region}
H.~Guo, G.~Wang, X.~Chen, C.~Zhang, F.~Qiao, and H.~Yang, ``Region ensemble
  network: Improving convolutional network for hand pose estimation,'' in
  \emph{Image Processing (ICIP), 2017 IEEE International Conference on}.\hskip
  1em plus 0.5em minus 0.4em\relax IEEE, 2017, pp. 4512--4516.

\bibitem{wang2018region}
G.~Wang, X.~Chen, H.~Guo, and C.~Zhang, ``Region ensemble network: Towards good
  practices for deep {3D} hand pose estimation,'' \emph{Journal of Visual
  Communication and Image Representation}, 2018.

\bibitem{chen2017pose}
X.~Chen, G.~Wang, H.~Guo, and C.~Zhang, ``Pose guided structured region
  ensemble network for cascaded hand pose estimation,'' \emph{Neurocomputing},
  2018.

\bibitem{ioffe2015batch}
S.~Ioffe and C.~Szegedy, ``Batch normalization: Accelerating deep network
  training by reducing internal covariate shift,'' in \emph{International
  Conference on Machine Learning}, 2015, pp. 448--456.

\bibitem{glorot2011deep}
X.~Glorot, A.~Bordes, and Y.~Bengio, ``Deep sparse rectifier neural networks,''
  in \emph{Proceedings of the Fourteenth International Conference on Artificial
  Intelligence and Statistics}, 2011, pp. 315--323.

\bibitem{leapmotion}
``{Leap Motion},'' {https://www.leapmotion.com/}.

\bibitem{girshick2015fast}
R.~Girshick, ``{Fast R-CNN},'' in \emph{Computer Vision (ICCV), 2015 IEEE
  International Conference on}.\hskip 1em plus 0.5em minus 0.4em\relax IEEE,
  2015, pp. 1440--1448.

\bibitem{trakstar}
``{Ascension Trakstar},'' {http://www.ascension-tech.com/}.

\bibitem{maaten2008visualizing}
L.~v.~d. Maaten and G.~Hinton, ``Visualizing data using {t-SNE},''
  \emph{Journal of machine learning research}, vol.~9, no. Nov, pp. 2579--2605,
  2008.

\bibitem{van2014accelerating}
L.~Van Der~Maaten, ``Accelerating {t-SNE} using tree-based algorithms.''
  \emph{Journal of machine learning research}, vol.~15, no.~1, pp. 3221--3245,
  2014.

\bibitem{jia2014caffe}
Y.~Jia, E.~Shelhamer, J.~Donahue, S.~Karayev, J.~Long, R.~Girshick,
  S.~Guadarrama, and T.~Darrell, ``Caffe: Convolutional architecture for fast
  feature embedding,'' in \emph{Proceedings of the 22nd ACM international
  conference on Multimedia}.\hskip 1em plus 0.5em minus 0.4em\relax ACM, 2014,
  pp. 675--678.

\end{thebibliography}
%



%

\newpage

\begin{IEEEbiography}[{\includegraphics[width=1in,height=1.25in,clip,keepaspectratio]{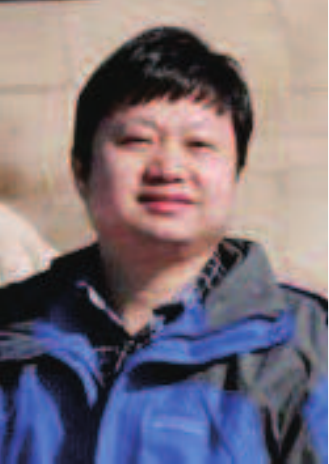}}]
{Guijin Wang} received the B.S. and Ph.D. degree (with honor) from the Department of Electronic Engineering, Tsinghua University, China in 1998 and 2003 respectively, all in signal and information processing. From 2003 to 2006, he has been with Sony Information Technologies Laboratories as a researcher. From Oct., 2006, he has been with the Department of Electronic Engineering, Tsinghua University, China as an Associate Professor. From 2012.01 to 2012.06, he was a visiting researcher in AMP Lab of Cornell. He won the reward (the first prize) of Science and Technology Award of Chinese Association for Artificial Intelligence in 2014, won the reward (the second prize) of Shandong Province Science and Technology Progress in 2014. He was Associate Editor of IEEE Signal Processing Magazine, the Guest Editor of Neurocomputing, the track chair of ChinaSIP 2015, the TPC member of ICIP2017. He published over 100 international journal and conference papers, holds tens of patents with numerous pending. His research interests focus on computational imaging, pose recognition, intelligent human-machine UI, intelligent surveillance, industry inspection, AI for Big medical data, etc.
\end{IEEEbiography}

\begin{IEEEbiography}[{\includegraphics[width=1in,height=1.25in,clip,keepaspectratio]{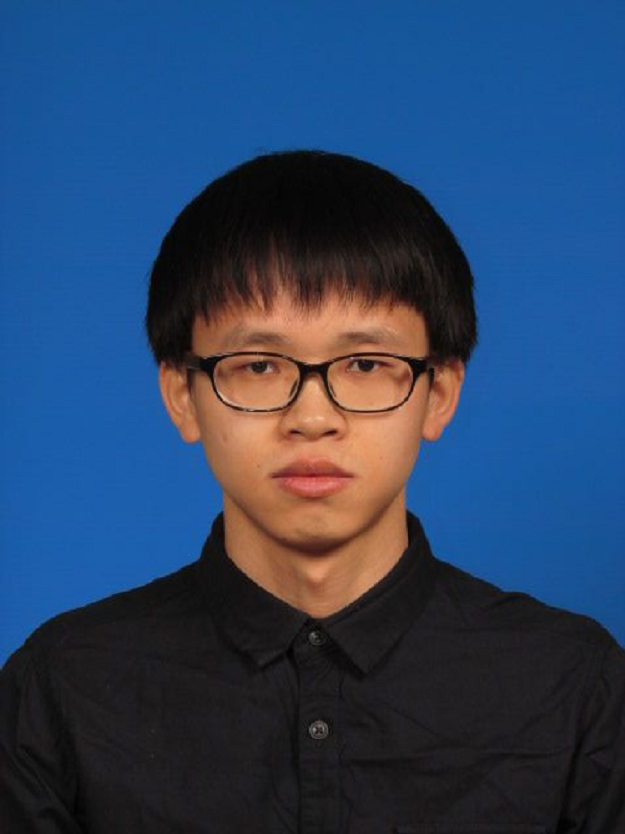}}]
{Cairong Zhang} received his B.S. degree from Department of Electronic Engineering, Tsinghua University, Beijing, China, in 2017, where he is currently working towards his M.S. degree. His research interests include deep learning, human pose estimation and hand pose estimation.
\end{IEEEbiography}

\begin{IEEEbiography}[{\includegraphics[width=1in,height=1.25in,clip,keepaspectratio]{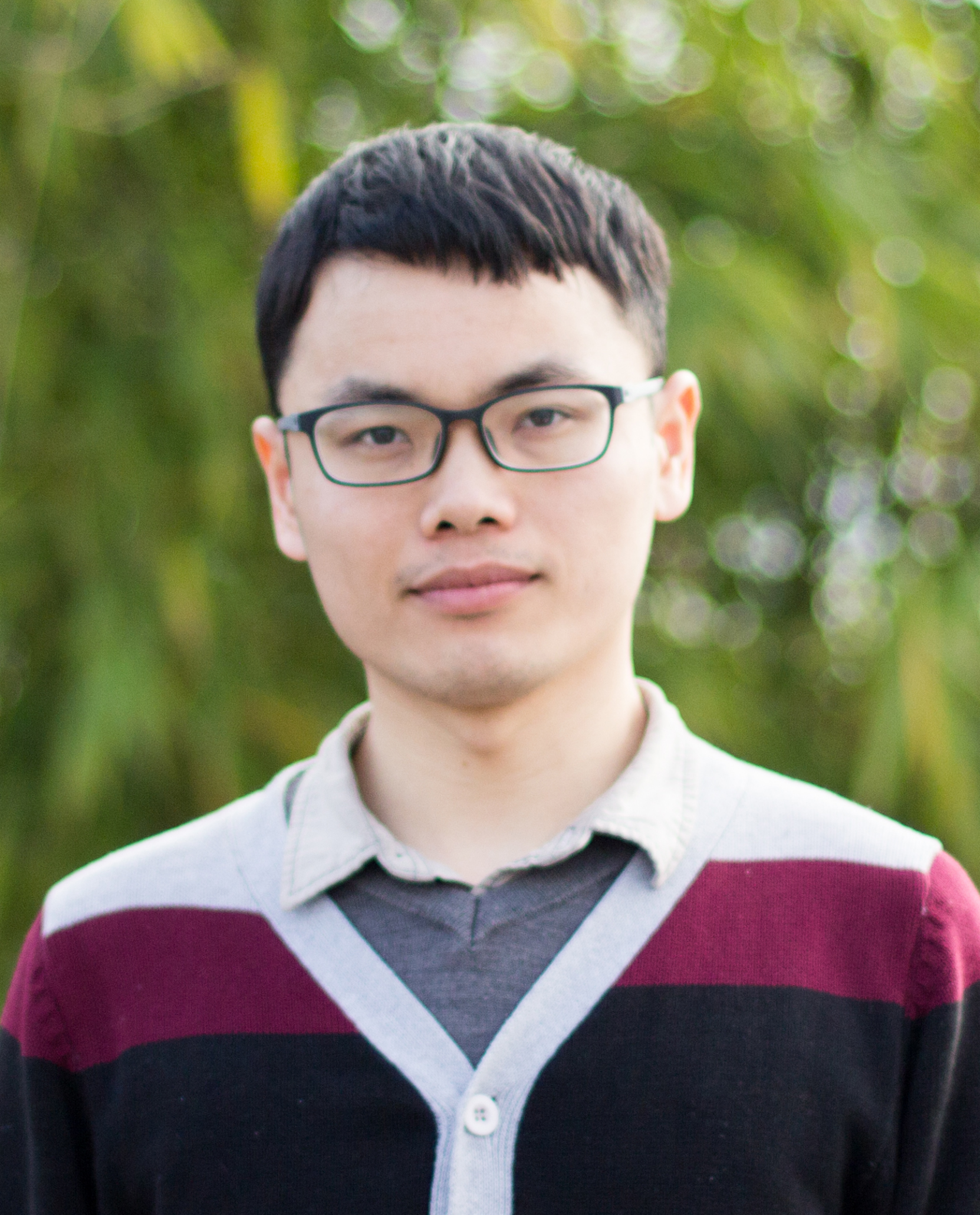}}]
{Xinghao Chen} received his B.S. and Ph.D. degree from Department of Electronic Engineering, Tsinghua University, Beijing, China, in 2013 and 2019 respectively. From Sept. 2016 - Jan. 2017, he was a visiting Ph.D. student with Imperial College London, UK. His research interests include deep learning, hand pose estimation and gesture recognition.
\end{IEEEbiography}

\newpage

\begin{IEEEbiography}[{\includegraphics[width=1in,height=1.25in,clip,keepaspectratio]{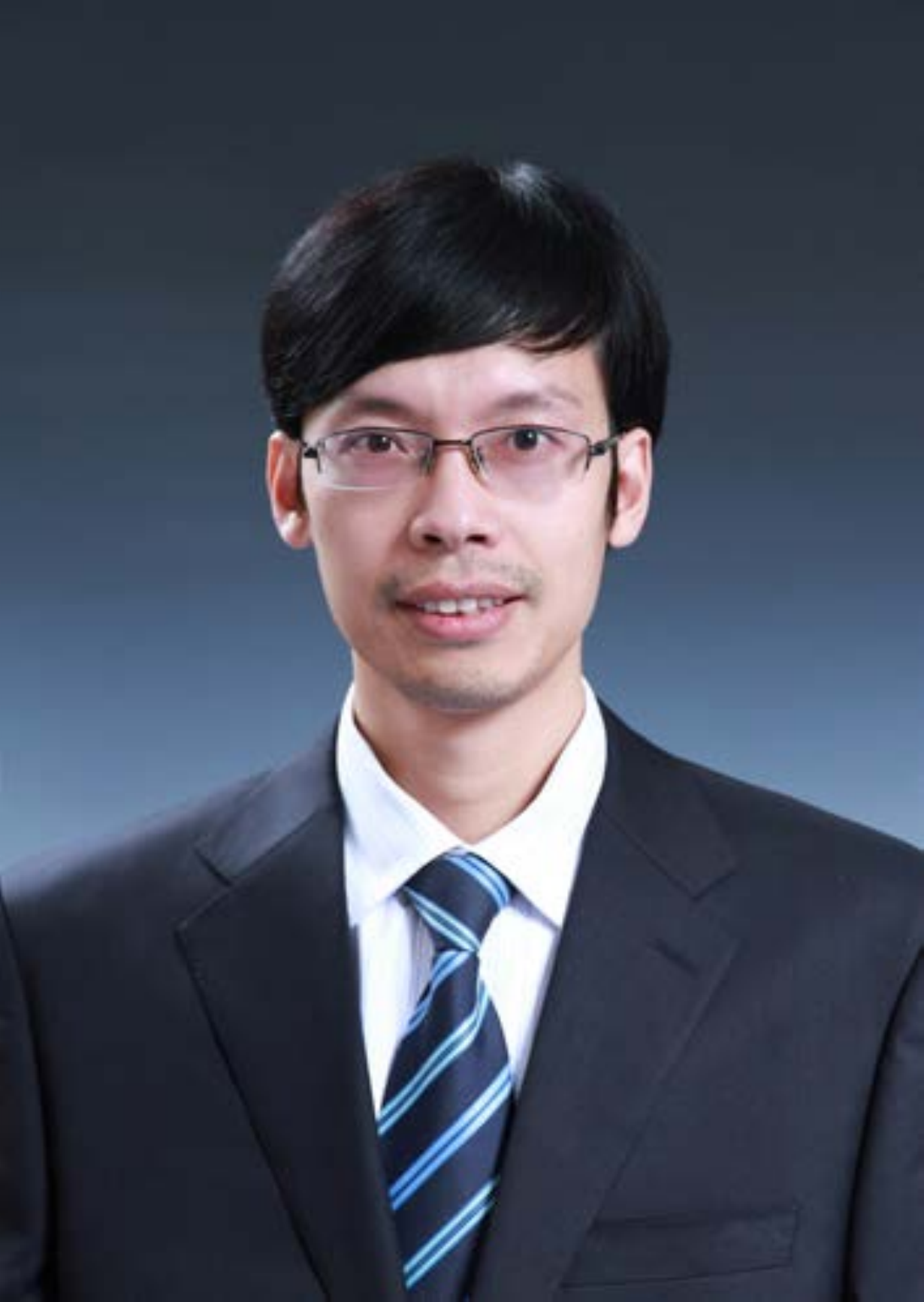}}]
{Xiangyang Ji} received the B.S. degree in materials science and the M.S. degree in computer science from the Harbin Institute of Technology, Harbin, China, in 1999 and 2001, respectively, and the Ph.D. degree in computer science from the Institute of Computing Technology, Chinese Academy of Sciences, Beijing, China. He joined Tsinghua University, Beijing, China, in 2008, where he is currently a Professor in the Department of Automation. His current research interests cover signal processing, image/video processing, machine learning.
\end{IEEEbiography}

\begin{IEEEbiography}[{\includegraphics[width=1in,height=1.25in,clip,keepaspectratio]{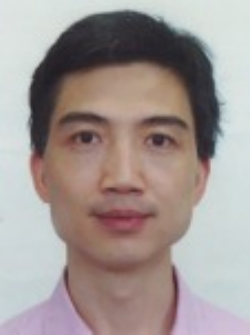}}]
{Jing-Hao Xue} received the Dr.Eng. degree in signal and information processing from Tsinghua University in 1998 and the Ph.D. degree in statistics from the University of Glasgow in 2008. He is a senior lecturer in the Department of Statistical Science, University College London. His research interests include statistical classification, high-dimensional data analysis, pattern recognition and image analysis.
\end{IEEEbiography}

\begin{IEEEbiography}[{\includegraphics[width=1in,height=1.25in,clip,keepaspectratio]{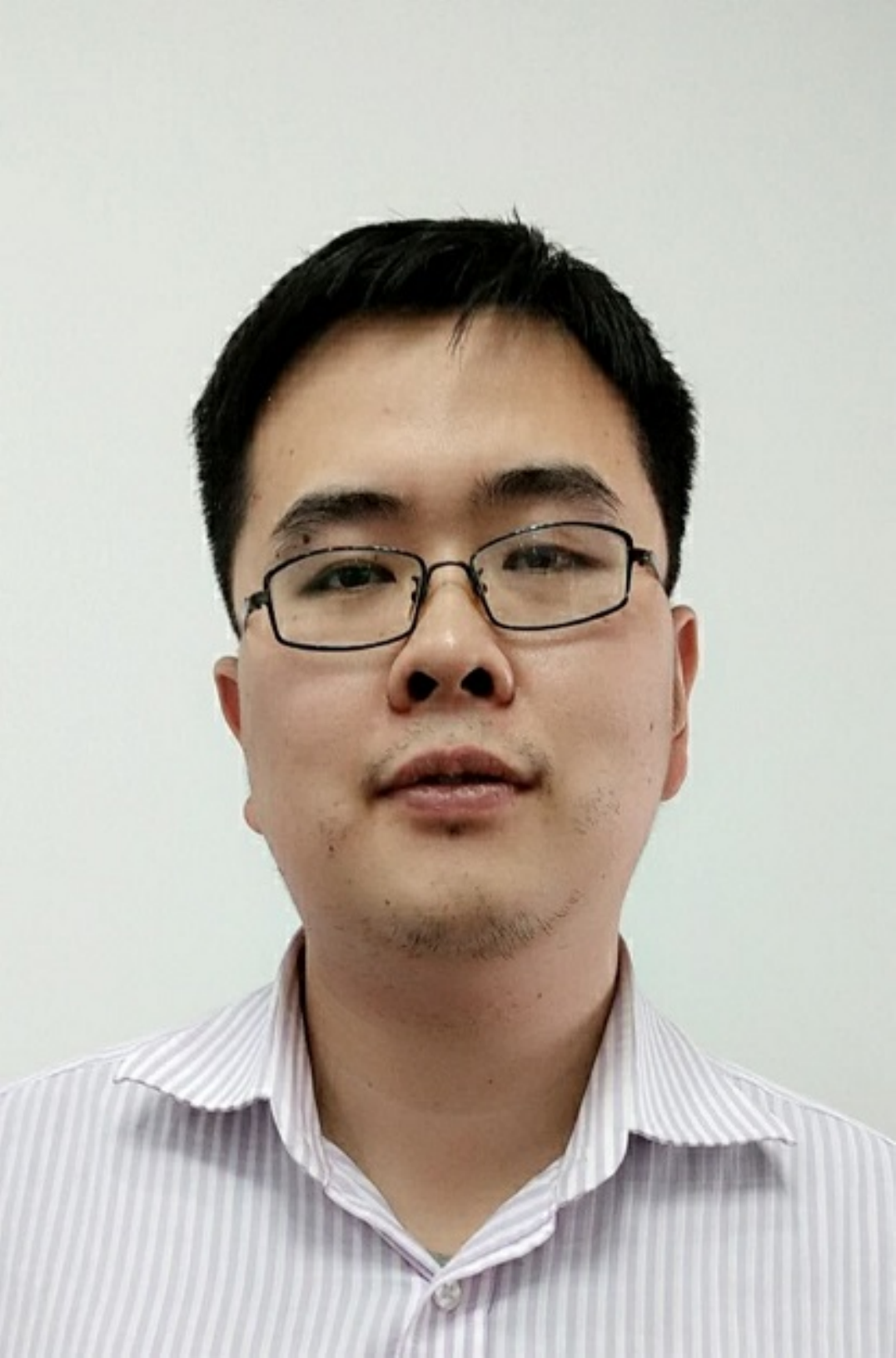}}]
{Hang Wang} received his B.S. degree in Automation from Beijing Institute of Technology in 2008. He received M.S degree in Control Science and Engineering from Beijing Institute of Technology in 2010. From 2014 to 2015, he was a senior software engineer in Baidu. He is now the vice president of Beijing HJIMI Technology Co.,Ltd since 2015.
\end{IEEEbiography}







\end{document}